\documentclass[conference]{IEEEtran}

\usepackage{cite}
\usepackage{amsmath,amssymb,amsfonts}
\usepackage{algorithm}
\usepackage{graphicx}
\usepackage{textcomp}
\usepackage{xcolor}
\usepackage{booktabs}
\usepackage{multirow}
\usepackage{url}
\usepackage{colortbl}
\usepackage{tikz}
\usetikzlibrary{shapes.geometric, arrows.meta, positioning, calc, fit}
\usepackage{tabularx}
\usepackage[table]{xcolor}
\usepackage{listings}
\usepackage{caption}
\usepackage{pifont}
\usepackage{booktabs}
\usepackage{multirow}
\usepackage{amsmath}
\usepackage{amssymb}
\usepackage{siunitx}
\DeclareSIUnit{\mps}{\metre\per\second}
\DeclareSIUnit{\kmph}{\kilo\metre\per\hour}

\usepackage{algorithm}
\usepackage{algpseudocode}

\usepackage{threeparttable}
\usepackage{wasysym}
\usepackage{array} 
\usepackage{makecell}

\definecolor{flashblue}{RGB}{33,150,243}
\definecolor{gpt4green}{RGB}{76,175,80}
\definecolor{kimiRed}{RGB}{255,87,34}
\definecolor{rowgray}{RGB}{245,245,245}
\definecolor{bestcell}{RGB}{232,245,233}
\definecolor{failcell}{RGB}{255,235,238}
\definecolor{headerblue}{RGB}{227,242,253}

\usepackage{tikz}
\usepackage{pgfplots}
\pgfplotsset{compat=1.18}
\usetikzlibrary{arrows.meta, calc, positioning, shapes.geometric}

\newtheorem{definition}{Definition}[section]

\begin{document}

\title{LIDSA: Cognitive Arbitration for Signal-Free Autonomous Intersection Management
}
\author{
    \IEEEauthorblockN{
    Abderrahmane~Lakas\IEEEauthorrefmark{1}\IEEEauthorrefmark{3},
    Mohamed~Amine~Ferrag\IEEEauthorrefmark{1},
    Merouane~Debbah\IEEEauthorrefmark{2}
    }
    \IEEEauthorblockA{\IEEEauthorrefmark{1}
    Department of Computer and Network Engineering,  
    United Arab Emirates University, UAE
    } 
    \IEEEauthorblockA{\IEEEauthorrefmark{2}
    Research Institute for Digital Future, Khalifa University, UAE
    }
    \IEEEauthorblockA{\IEEEauthorrefmark{3}
    Corresponding author: \texttt{alakas@uaeu.ac.ae}
    }
}

\maketitle

\begin{abstract}

Large language models (LLM) show strong potential for Intelligent Transportation Systems (ITS), especially for tasks requiring situational reasoning and multi-agent coordination. This makes them well suited to cooperative driving, where rule-based methods often struggle with complex and dynamic traffic conditions.
Intersection management is a particularly challenging setting, as it must resolve conflicting right-of-way demands, heterogeneous vehicle priorities, and vehicle-specific kinematic constraints in real time. However, existing approaches typically use LLMs as auxiliary reasoning modules on top of conventional signal-control systems, rather than as autonomous agents for direct intersection-level decision-making. Signal-based controllers remain vehicle-agnostic, reservation-based methods lack intent awareness, and recent LLM-based systems still depend on signal infrastructure. Moreover, LLM inference latency further limits their use, since sub-second control cycles cannot wait for multi-second reasoning calls. In this paper, we propose LIDSA (LLM-based Intent-Driven Speed Advisory), a signal-free cognitive arbitration method for autonomous intersection management. LIDSA uses an LLM to arbitrate declared vehicle intents by reasoning over priority classes, queue pressure, and energy preferences. We evaluate LIDSA against conventional intersection management methods, including fixed-cycle control, SCATS, AIM, and GLOSA, across three traffic load levels. We show that our approach reduces mean control delay by up to {89.1\%}, maintains {Level of Service~C} while all non-LLM baselines degrade to {Level of Service~F}, and, under near-saturated demand, reduces mean waiting time by {93\%} and peak queue length by {60.6\%} relative to fixed-cycle control. LIDSA further lowers fuel consumption by up to 48.8\% relative to fixed-cycle
control and achieves an overall intent satisfaction of 86.2\%, compared with
the best non-LLM result of 61.2\%. These results demonstrate that LLM reasoning, combined with lightweight latency mitigation, can support real-time cooperative intersection arbitration without signal infrastructure.

\end{abstract}


\maketitle

\section{Introduction}
\label{sec:introduction}

Urban transportation systems have historically been designed around human-operated vehicles that require external physical cues (traffic lights, signs, and signal-phase controllers) to negotiate shared road
space safely~\cite{zhao2011computational}. This
infrastructure-centric paradigm has shaped Intelligent Transportation Systems (ITS) for decades, but it also imposes a structural limitation: traffic flow is regulated through aggregate, reactive commands rather
than through direct reasoning over the intentions and capabilities of individual vehicles~\cite{zhu2019parallel}. The emergence of Connected and Autonomous Vehicles (CAVs), equipped with onboard sensing, cooperative perception, and Vehicle-to-Everything (V2X) communication, challenges this assumption~\cite{ahangar2021survey}. If vehicles are able to perceive their surroundings, exchange information, and follow coordinated advisories, then intersection management no longer needs to be strictly tied to physical signal phases originally designed for human drivers.

This shift is especially important at urban intersections, where vehicles from competing approaches must share the same conflict zone while avoiding collisions, deadlock, and unnecessary delay. Traditionally, this responsibility has been assigned to traffic light infrastructure~\cite{chen2016cooperative,namazi2019aim}. In a CAV-populated environment, however, vehicles can communicate their intended movements, priority classes, and operating preferences in advance. What is needed is a coordination mechanism capable of interpreting these intents and converting them into safe, conflict-free crossing decisions in real time. From this perspective, intersection management can be reformulated as a \textit{cognitive arbitration} problem, in which a reasoning agent jointly interprets the crossing intents of all active approaches and assigns cooperative right-of-way roles before vehicles enter the conflict zone.

Existing approaches address this problem only partially, each bounded
by a distinct structural limitation. Signal-based control, including fixed-cycle systems~\cite{albdairi2026webster}
and adaptive controllers such as Max Pressure variants~\cite{riehl2025greenpressure}, remains fundamentally vehicle-agnostic:
phases are derived from aggregate flow or occupancy measurements rather
than from vehicle-level intent, priority, or route information~\cite{zhu2019parallel,eom2020traffic,qadri2020state}.
Green Light Optimal Speed Advisory (GLOSA) improves kinematic efficiency
by translating Signal Phase and Timing (SPaT) information into
approach-speed recommendations~\cite{stevanovic2013green,eckhoff2013potentials}, yet it optimizes individual
vehicle motion within an existing phase schedule and provides no
mechanism for multi-vehicle right-of-way resolution~\cite{bhattacharyya2022assessing}.
Autonomous Intersection Management (AIM) eliminates the signal entirely
by assigning reservation slots or conflict-zone tiles~\cite{dresner2008aim},
but coordination remains deterministic scheduling over vehicle kinematic
state, with no capacity to reason over crossing intent, priority class,
queue pressure, or contextual trade-offs~\cite{namazi2019aim,
khayatian2020survey,gholamhosseinian2022survey,wu2019dcl}. What is
absent across all three paradigms is a controller that can read the
situation, one that understands not just where vehicles are, but
what they intend, and can arbitrate accordingly.

Large Language Models (LLMs) offer a qualitatively different foundation
for this problem~\cite{ferrag2026llm}. Their capacity for open-vocabulary reasoning,
contextual inference, and multi-party deliberation maps naturally onto
the structure of intersection negotiation, where competing claims over
a shared conflict zone must be resolved under heterogeneous priorities
and real-time constraints~\cite{tian2025large,jing2026multi}. Recent
work has begun to explore this alignment: LLM-based agents have been
shown to select signal phases through chain-of-thought reasoning over
queue and delay state~\cite{movahedi2025crossroads,lai2025llmlight},
to audit and correct RL decisions under degraded communication or
emergency conditions~\cite{pang2026illmtsc}, and to coordinate across
multiple intersections by supplying predictive context to local
agents~\cite{zhu2025llmag}. The most semantically expressive of these
formulations processes spatial, temporal, contextual, and natural
language intent simultaneously to generate conflict-aware
decisions~\cite{masri2025llm}. Yet in each case the LLM operates
within or alongside a signal-phase actuation layer: it reasons about
traffic, but it does not replace the infrastructure that traffic was
designed around. The question that remains open is whether LLM
reasoning alone, without signal phases, without reservation
schedules, and without task-specific training, is sufficient to
arbitrate crossing intent and deliver cooperative kinematic advisories
to approaching vehicles.

To address this gap, we introduce LIDSA (LLM-Based Intent-Driven Speed Advisory), an cognitive arbitration framework for signal-free autonomous intersection management. Unlike existing methods, LIDSA resolves right-of-way through intent-based arbitration rather than aggregate flow measurements or
deterministic schedules. It overcomes LLM latency constraints through
two complementary mechanisms: a Memoized Arbitration Table (MAT) for recurring
conflict signatures and anticipatory arbitration triggered before
vehicles reach the conflict zone. The large language model functions
not as a signal optimizer but as an explicit arbitration agent, reasoning over both vehicle intent and the traffic situation around the conflict zone. Each arbitration outcome is then passed to a deterministic kinematic executor, which converts the decision into vehicle-specific speed advisories. This design preserves the interpretability of the LLM’s reasoning while providing the temporal precision required for vehicle-level control.

The main contributions of this paper are as follows:

\begin{figure}[!t]
    \centering
    \includegraphics[width=1\linewidth]{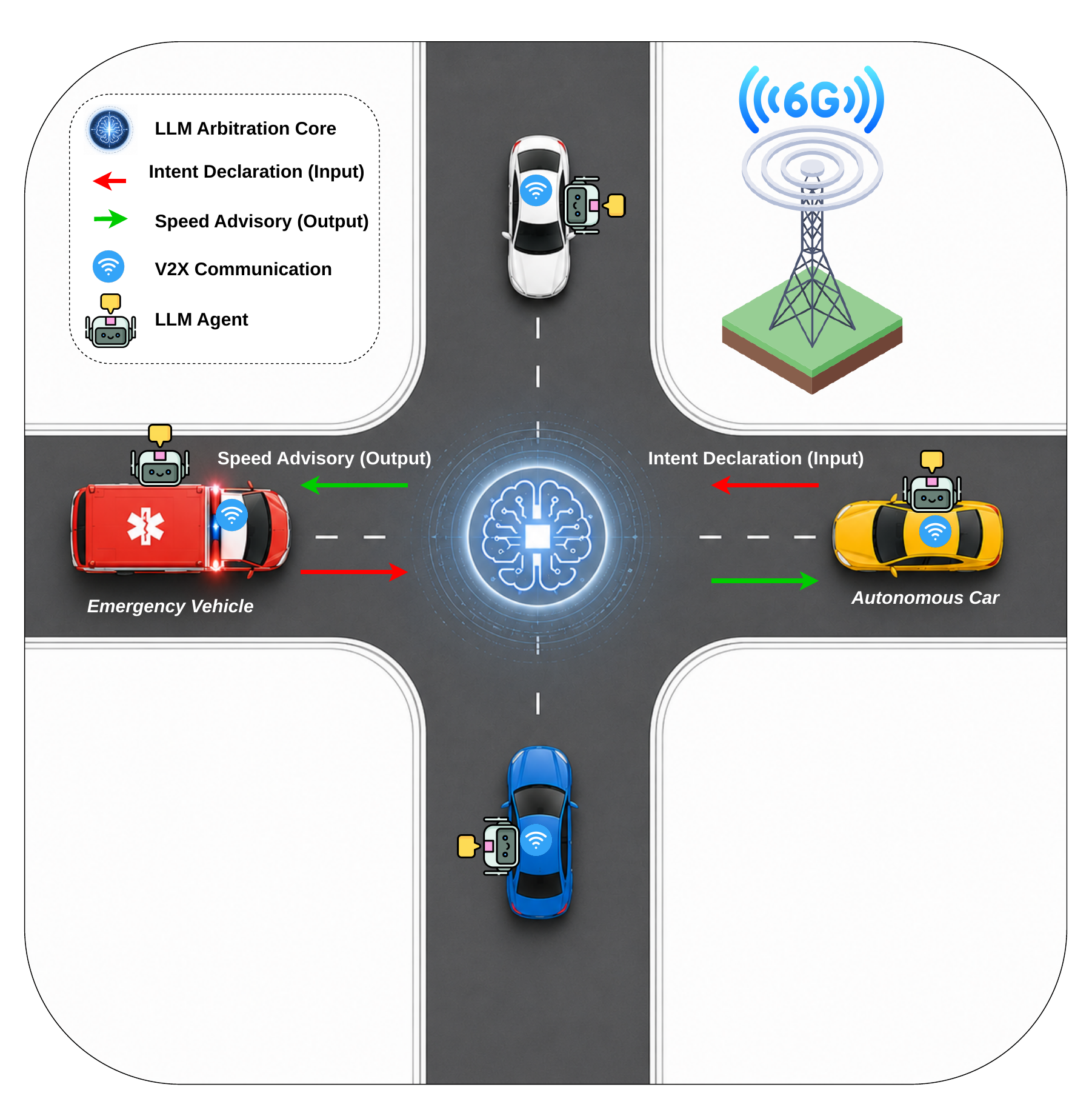}
    \caption{Free-signal Intersection Management using LLM-based Arbitration.}
    \label{fig:llm_arbitration}
\end{figure}

\begin{enumerate}

  \item \textit{Signal-free intent-based arbitration framework}: 
  We formulate autonomous intersection management as a cognitive
  arbitration problem and propose LIDSA, a signal-free architecture
  in which an LLM reasoning agent resolves competing right-of-way
  claims directly from approach-level vehicle intents, without signal
  phases, reservation schedules, or task-specific training.

  \item \textit{Two-layer control architecture with latency mitigation:}
  We design a separation between high-level semantic arbitration and
  low-level kinematic execution, in which the LLM resolves right-of-way
  at the intent level while a deterministic executor maps each decision
  to verifiable per-vehicle speed advisories. To overcome LLM latency
  constraints at intersection timescales, LIDSA employs two complementary mechanisms: recurring arbitration
  signatures are cached using MAT, which achieves hit rates of up to {98.8\%} at
  medium demand, and anticipatory arbitration triggered before vehicles
  reach the conflict zone, sustaining zero LLM fallbacks across all
  evaluated conditions.

  \item \textit{Empirical evaluation across demand levels and LLM backends:}
  We evaluate LIDSA in SUMO under low, medium, and near-saturated load
  conditions against fixed-cycle control,
  SCATS adaptive control, AIM, and GLOSA, using Gemini-2.5 Flash Lite as LLM backends. LIDSA achieves the highest throughput and lowest mean control delay across all load levels, and reduces average fuel consumption by up to {51\%} and kinetic energy
  loss by up to {26\%} relative to fixed-cycle control at
  medium demand.

  \item \textit{Multi-dimensional intent satisfaction metric:}
  We introduce a structured evaluation axis decomposing intersection
  performance into spatial, temporal, priority, and energy satisfaction
  components. Under this metric, LIDSA achieves an overall intent
  satisfaction of {86.2\%} at medium demand against the best performance of {61.2\%} for non-LLM methods, demonstrating that cognitive arbitration yields measurable gains across all intent dimensions simultaneously.

\end{enumerate}

The remainder of this paper is organized as follows.
Section~\ref{sec:related} reviews related work on signal-based control,
GLOSA and V2I advisory systems, autonomous intersection management, and
LLM-based traffic reasoning. Section~\ref{sec:system_design} presents
the LIDSA architecture, including the intent representation, role-based
arbitration protocol, speed-mapping layer, cache, and safety watchdog.
Section~\ref{sec:llm_benchmark} reports the LLM policy orchestration
benchmark used to select the deployment model. Section~\ref{sec:setup}
describes the SUMO simulation environment, demand scenarios, baselines,
and evaluation metrics. Section~\ref{sec:performance} presents the
experimental results. Section~\ref{sec:discussion} discusses the main
findings and limitations, and Section~\ref{sec:conclusion} concludes
the paper.

\section{Related Work}
\label{sec:related}

Intersection management has evolved through four broad paradigms:
fixed-time and actuated signal control, adaptive signal control,
autonomous intersection management (AIM), and, most recently,
LLM-based traffic reasoning. GLOSA and V2I speed advisory systems
occupy an intermediate position, improving vehicle-level efficiency
within the signal-infrastructure paradigm without eliminating it.
This section reviews representative work in each category and
situates LIDSA within the resulting landscape.

\begin{table*}[ht]
\centering
\scriptsize
\caption{Compact Comparison of Intersection Management Paradigms}
\label{tab:full_comparison}
\begin{threeparttable}
\renewcommand{\arraystretch}{1.15}
\setlength{\tabcolsep}{4pt}
\begin{tabular}{@{}p{5cm}cccccccp{3.0cm}@{}}
\toprule
\textbf{Method / Paradigm}
& \textbf{\makecell{Signal-\\Free}}
& \textbf{\makecell{Intent\\Aware}}
& \textbf{V2X}
& \textbf{\makecell{Speed\\Advisory}}
& \textbf{\makecell{Conflict\\Resolution}}
& \textbf{\makecell{Vehicle\\Aware}}
& \textbf{Expl.}
& \textbf{Control Paradigm} \\
\midrule

Fixed-time/actuated TSC~\cite{duraku2024fixed,albdairi2026webster,eom2020traffic}
& \Circle & \Circle & \LEFTcircle & \Circle & \LEFTcircle & \Circle & \Circle
& Phase-based control \\

Adaptive/learning-based TSC~\cite{chen2016cooperative,wu2024deeprl,shafik2025gametheoretic,tan2024genetic}
& \Circle & \Circle & \LEFTcircle & \Circle & \LEFTcircle & \Circle & \Circle
& Adaptive phase selection \\

GLOSA/SPaT advisory~\cite{eckhoff2013potentials,stevanovic2013green}
& \Circle & \Circle & \CIRCLE & \CIRCLE & \Circle & \LEFTcircle & \Circle
& Signal-based speed advisory \\

Reservation-based AIM~\cite{dresner2008aim,guan2025decentralized,rostomyan2025central,pan2025cooperative}
& \CIRCLE & \LEFTcircle & \CIRCLE & \Circle & \CIRCLE & \LEFTcircle & \Circle
& Slot / tile reservation \\

LLM-based signal control~\cite{lai2025llmlight,movahedi2025crossroads,pang2026illmtsc,zhu2025llmag}
& \Circle & \LEFTcircle & \LEFTcircle & \Circle & \LEFTcircle & \Circle & \LEFTcircle\tnote{1}
& LLM-assisted phase control \\

Semantic LLM control~\cite{masri2025llm}
& \Circle & \CIRCLE & \LEFTcircle & \CIRCLE & \CIRCLE & \Circle & \LEFTcircle
& CoT-based signal control \\

\textbf{LIDSA (Proposed)}
& \CIRCLE & \CIRCLE & \CIRCLE & \CIRCLE & \CIRCLE & \CIRCLE & \CIRCLE\tnote{2}
& \textbf{Zero-shot intent arbitration} \\

\bottomrule
\end{tabular}

\begin{tablenotes}
\scriptsize
\item \textbf{Symbols:}
\CIRCLE~=~fully supported;\quad
\LEFTcircle~=~partially supported;\quad
\Circle~=~not supported.
\item\tnote{1} Existing LLM-based traffic controllers mainly explain signal-phase choices rather than explicit right-of-way roles.
\item\tnote{2} LIDSA explains conflict resolution and right-of-way assignments directly through structured intent-based arbitration.
\end{tablenotes}

\end{threeparttable}
\end{table*}

\subsection{Signal-Based and Adaptive Control}
\label{sec:related:signal_adaptive}

Traffic signal control remains the dominant mechanism for urban intersections. Fixed-time, actuated, and adaptive controllers improve queue length, travel time, and throughput by adjusting phase duration, cycle length, or phase sequence in response to traffic observations~\cite{eom2020traffic,qadri2020state,agrahari2024artificial,hassan2026comprehensive}. However, these methods remain phase-based and primarily operate on aggregate flow or occupancy measurements rather than vehicle-level intent, priority, or route information.

\subsection{GLOSA and V2I Speed Advisory Systems}
\label{sec:related:glosa}

Green Light Optimal Speed Advisory (GLOSA) systems~\cite{eckhoff2013potentials} mark an important step from purely infrastructure-centered signal control toward vehicle-aware guidance. Rather than changing the signal plan itself, GLOSA uses Signal Phase and Timing (SPaT) information to recommend approach speeds that allow vehicles to pass intersections more smoothly. These advisories can reduce unnecessary stops, moderate acceleration and deceleration, and improve fuel or energy efficiency~\cite{stevanovic2013green,eckhoff2013potentials}. Empirical studies have shown that GLOSA can improve both traffic performance and environmental outcomes, provided that signal timing information is reliable and drivers or automated vehicles follow the recommended speeds appropriately~\cite{bhattacharyya2022assessing}. Vehicle-to-infrastructure (V2I) and cellular vehicle-to-everything (C-V2X) communication standards provide the connectivity layer through which SPaT messages and GLOSA advisories are delivered~\cite{alalewi2021v2x}.

GLOSA, however, remains architecturally tied to traffic signals. Its advisory speeds are computed from an existing phase schedule rather than from direct arbitration among vehicles. As a result, its performance depends on the accuracy of signal timing information, market penetration~\cite{bhattacharyya2022assessing}, and vehicle compliance with the recommended speeds~\cite{eckhoff2013potentials}.  In contrast, LIDSA uses structured vehicle intent to infer speed advisories from LLM-based arbitration rather than signal phases.

\subsection{Autonomous Intersection Management}
\label{sec:related:aim}

Autonomous Intersection Management (AIM), introduced by Dresner and Stone~\cite{dresner2008aim}, was among the first signal-free approaches to intersection control. It replaces phase-based right-of-way with a reservation protocol, where a central arbiter grants each vehicle a time-space slot in the conflict zone. The first-come-first-served (FCFS) policy has served as the basis for many extensions, including learning-based and decentralized coordination methods that address the limitations of simple reservation rules~\cite{wu2019dcl}. Recent work has further extended AIM through decentralized CAV negotiation using V2V and V2I communication~\cite{guan2025decentralized}, centralized slot assignment with formal safety guarantees~\cite{rostomyan2025central}, and distributed trajectory planning for networks of unsignalized intersections~\cite{pan2025cooperative}. However, AIM-based methods do not natively reason about intent, priority, or contextual trade-offs~\cite{zhong2021semiautonomous,namazi2019aim}. LIDSA shares AIM's signal-free design but replaces reservation-based control with semantic right-of-way assignment, enabling intent-aware arbitration while preserving deterministic execution and rule-based safety verification.

\subsection{LLM-Based Intersection Control}
\label{sec:related:llm}

The application of large language models to traffic control is recent
but rapidly expanding. Tian et al.~\cite{tian2025large} and Jing
et al.~\cite{jing2026multi} survey the broader role of LLMs and
vision-language models in autonomous driving, identifying scene
understanding, planning, and decision explanation as the primary
capability. Within intersection management specifically, several
architectures have been proposed that differ in their degree of
signal-dependence, reasoning modality, and infrastructure requirements.

LLMLight~\cite{lai2025llmlight} encodes real-time intersection state into structured natural-language prompts and uses an LLM to select the next signal phase via chain-of-thought reasoning; a fine-tuned variant, LightGPT, matches state-of-the-art RL performance while generalizing better to unseen scenarios. Movahedi and Choi~\cite{movahedi2025crossroads} similarly deploy LLMs as adaptive signal controllers, with their CGA (Generally Capable Agent) reducing halted vehicle counts by 48.03\% and average speed by 25.29\% over conventional baselines. Both works position the LLM as a signal-phase selector rather than a signal-free arbitrator.
Pang et al.~\cite{pang2026illmtsc} propose iLLM-TSC, which combines
reinforcement learning with an LLM correction layer. An RL agent
generates an initial phase decision; the LLM then evaluates and
optionally overrides it under special conditions such as degraded
communication or unmodeled emergency vehicles. Zhu and
Zhu~\cite{zhu2025llmag} introduce LLM-AG, a hierarchical multi-agent
framework in which a centralized LLM agent provides predictive
traffic-inflow information to local RL-based intersection agents,
enriching their state representations and improving global
coordination. In both architectures, the LLM acts as a supervisory
corrector rather than a primary arbitrator.

Masri et al.~\cite{masri2025llm} propose a 4D-LLM controller that uses CoT prompting to reason across four intent dimensions (spatial, temporal, contextual, and semantic) replacing the traditional signal control pipeline (detection, prediction, and optimization stages) with a single LLM-driven decision process. The 4D-LLM controller is the closest prior work to LIDSA in its use of semantic vehicle intent and structured CoT deliberation; however, it operates within a signalized intersection context, and exposes neither a deterministic speed-advisory executor nor a formal right-of-way-assignment protocol.

\subsection{Positioning of LIDSA}
\label{sec:related:positioning}

The reviewed literature reveals a clear trajectory toward more
connected, adaptive, and cooperative intersection management.
Non-LLM methods progressively advance from fixed signal phases
toward vehicle-level adaptivity through speed advisory systems,
reservation-based protocols, and learning-based control, while
LLM-based methods introduce open-vocabulary reasoning and
semantic awareness but largely preserve signal-phase logic,
selecting phases or adjusting timing rather than replacing the
signal infrastructure itself.

LIDSA addresses the gap left by both families of methods through three design choices that together distinguish it from all prior work. First, it formulates intersection management as \textit{cognitive arbitration}: the LLM jointly interprets approach-level intents and generates right-of-way assignments before vehicles enter the conflict zone, providing explicit and human-readable conflict resolution. Second, it separates high-level arbitration from low-level control through a deterministic kinematic executor that maps each assignment to a verifiable speed advisory using live distance-to-stop-line measurements, removing the LLM from the real-time control loop. Third, it operates in a zero-shot regime without fine-tuning or simulation-driven training, relying on the LLM's general reasoning capability augmented by a memoized arbitration table (MAT) and an independent tile-based safety watchdog.

Table~\ref{tab:full_comparison} summarizes the reviewed methods across the criteria discussed above, highlighting the progressive shift from signal-dependent to signal-free and intent-aware control.

\begin{figure*}[t]
    \centering
    \includegraphics[width=\textwidth]{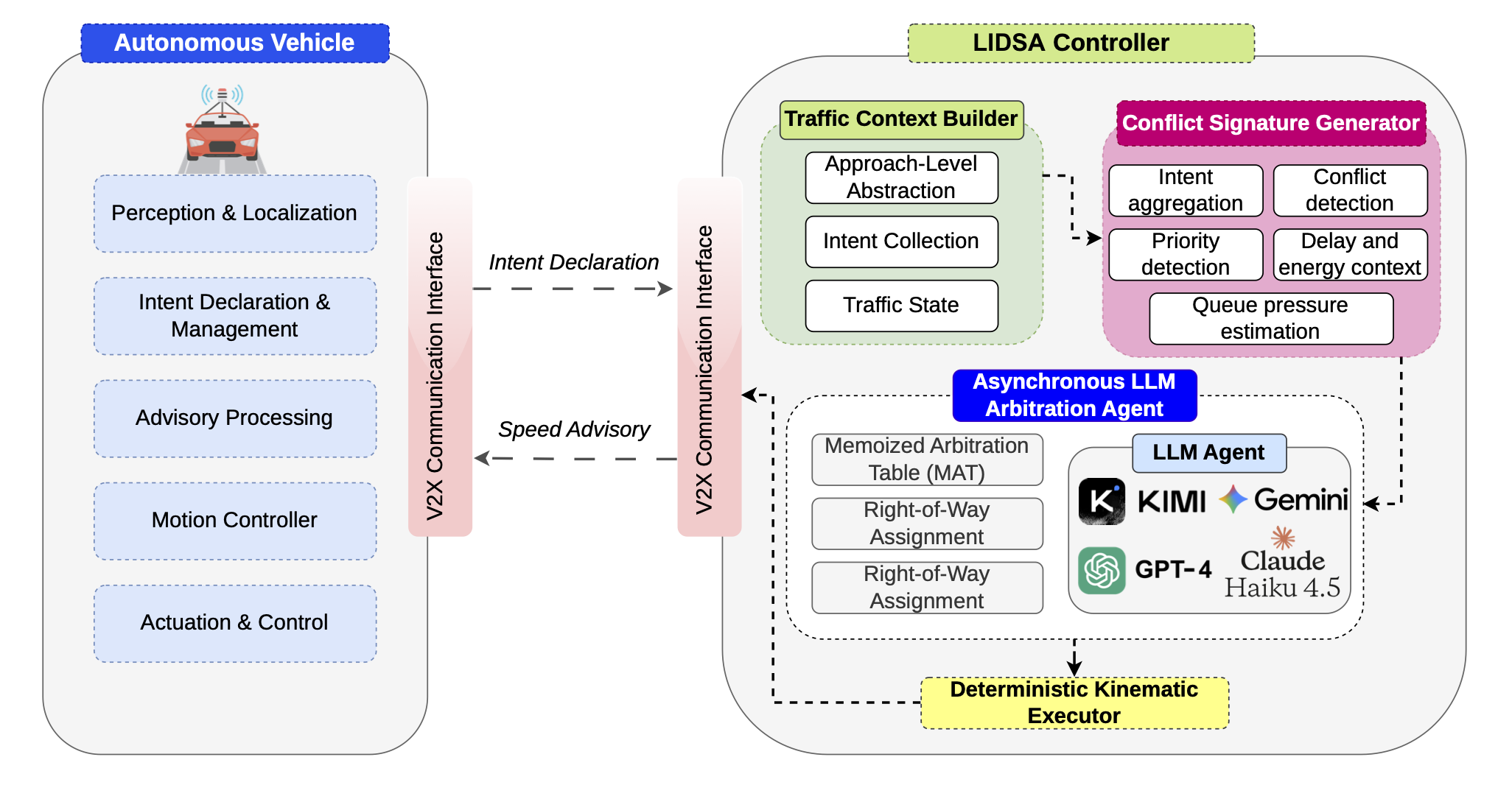}
\caption{LIDSA architecture for signal-free, intent-based intersection management, combining LLM-driven arbitration, cache-based optimization, and deterministic execution in a closed-loop control system.}
    \label{fig:LIDSA_architecture}
\end{figure*}

\section{System Design}
\label{sec:system_design}

LIDSA frames signal-free intersection management as a cognitive
right-of-way arbitration problem. Instead of enforcing movement through
fixed signal phases, an LLM reasoning agent interprets the declared
intents of approaching vehicles and assigns approach-level
right-of-way roles before vehicles enter the conflict zone
(Fig.~\ref{fig:arbitration_zone}). A deterministic kinematic executor
then converts these roles into per-vehicle speed advisories. This
separation isolates high-level conflict reasoning from low-level speed
control.

\begin{definition}[Cognitive arbitration]
Cognitive arbitration is the process of mapping a structured description of
active approaches, declared vehicle intents, priority classes, queue pressures,
and conflict relations to symbolic right-of-way roles.
\end{definition}

\subsection{Signal-Free Operating Principle}
\label{subsec:operating_principle}

LIDSA keeps the crossing zone continuously open: there are no signal
phases, red periods, or stop-and-wait cycles. Instead, right-of-way is
resolved through cooperative speed advisories. Delay is introduced only
when a geometric conflict requires one approach to defer to another.

\begin{figure}[t]
    \centering
    \includegraphics[width=.8\linewidth]{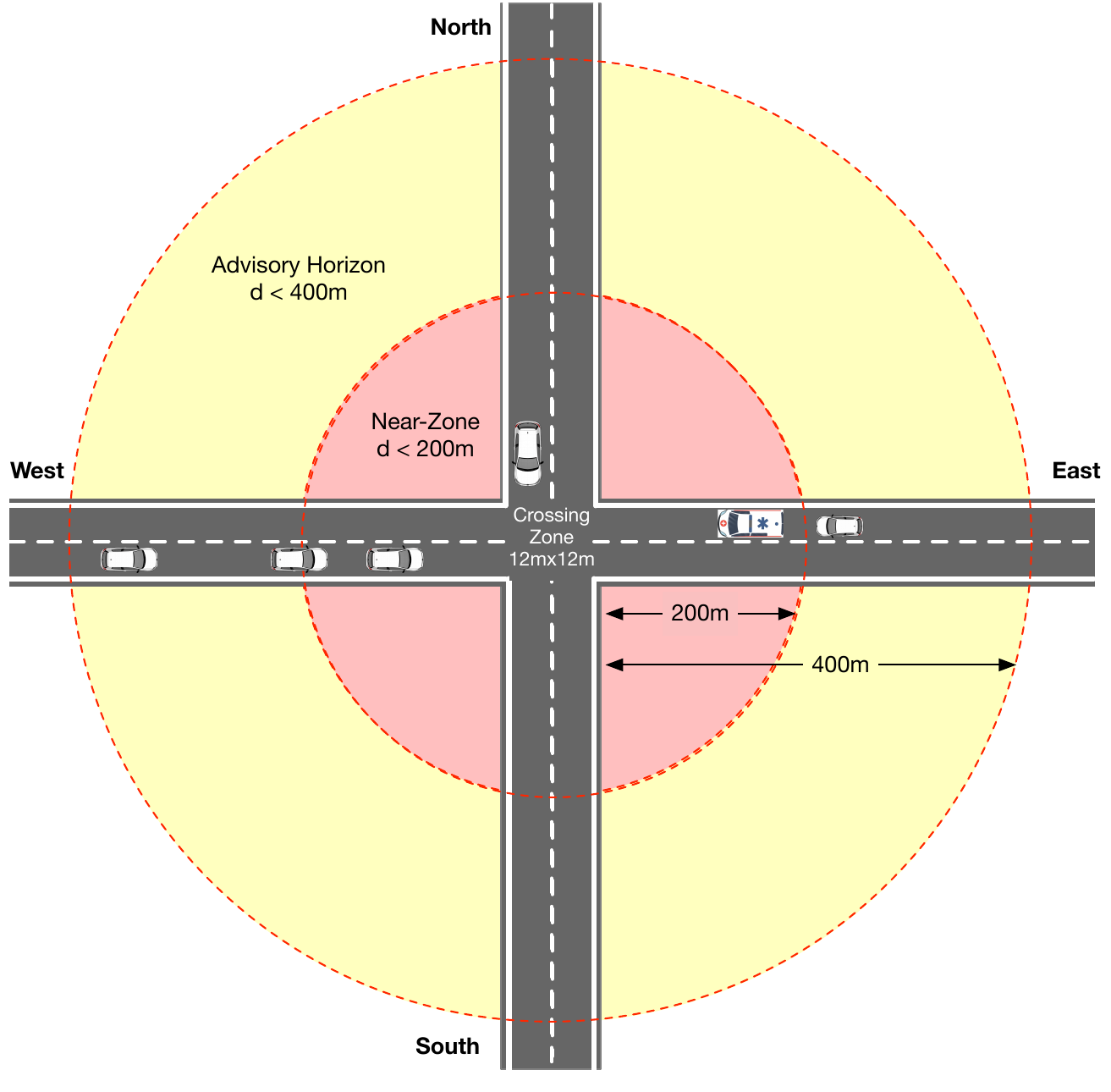}
    \caption{A signal-free four-way intersection with the advisory and near-zone regions.}
    \label{fig:arbitration_zone}
\end{figure}

The approach corridor is divided into two regions relative to the stop
line. The \textit{advisory horizon} ($d \leq \SI{400}{\metre}$) marks
the point at which a vehicle becomes visible to the arbitration agent.
The \textit{near zone} ($d \leq \SI{200}{\metre}$) is the region in
which assigned right-of-way roles are actively enforced through speed
advisories, as shown in Fig.~\ref{fig:arbitration_zone}. Vehicles
outside the advisory horizon travel at free-flow speed $V_{\max}$.

Queries are issued when vehicles enter the advisory horizon rather than
when they reach the near zone. This creates a latency buffer of
\begin{equation}
  t_{\mathrm{buffer}} =
  \frac{d_{\mathrm{horizon}} - d_{\mathrm{near}}}{V_{\max}}
  =
  \frac{400 - 200}{13.89}
  \approx \SI{14}{\second},
  \label{eq:latency_buffer}
\end{equation}
allowing LLM responses to arrive before the vehicle reaches the region
where advisory speeds are enforced.

LLM queries are dispatched asynchronously in a non-blocking background
thread. The 1\,Hz control loop polls for completed responses at each
step; if no new response is available, the most recent valid assignment
from the Memoized Arbitration Table (MAT) remains in effect. Advisory
speeds are recomputed at every control step from the cached assignment
and the current vehicle distance, so LLM inference latency never stalls
the control loop.

To bound prompt size, the LLM receives one entry per active approach,
corresponding to the leading vehicle nearest the stop line. Non-leading
vehicles are handled deterministically by the execution layer using the
\textsc{Follow} rule in Section~\ref{subsec:speed_mapping}. Thus, the
LLM prompt contains at most four approach entries regardless of queue
depth.

\subsection{Intersection Model and Notation}
\label{subsec:intersection_model}

We model the road geometry as a single four-way intersection with four
inbound approaches,
\begin{equation}
  \mathcal{A} = \{N, E, S, W\},
  \label{eq:approach_set}
\end{equation}
corresponding to the \textit{north}, \textit{east}, \textit{south}, and \textit{west} entries into the
crossing zone. At time $t$, the set of active approaches is
denoted by $\mathcal{A}_t \subseteq \mathcal{A}$. An approach is active
if at least one vehicle on that approach lies within the \textit{advisory
horizon}.
Table~\ref{tab:intersection_notation} summarizes the notation used in
this section.

\begin{table}[!t]
\centering
\caption{Main notation used in the LIDSA framework.}
\label{tab:intersection_notation}
\renewcommand{\arraystretch}{1.12}
\setlength{\tabcolsep}{1pt}
\begin{tabular}{p{0.20\columnwidth} p{0.7\columnwidth}}
\toprule
\textbf{Symbol} & \textbf{Description} \\
\midrule

\ensuremath{\mathcal{A}} &
Set of all inbound approaches to the intersection. \\

\ensuremath{t} &
Current simulation time or control step. \\

\ensuremath{\mathcal{A}_t} &
Set of active approaches at time \ensuremath{t}. \\

\ensuremath{a,b} &
Generic approaches. When \ensuremath{a,b \in \mathcal{A}_t}, they denote
active approaches evaluated for conflict. \\

\ensuremath{i,j} &
Vehicle indices. \\

\ensuremath{a_i} &
Approach from which vehicle \ensuremath{i} enters the intersection. \\

\ensuremath{\ell_a(t)} &
Leading vehicle on approach \ensuremath{a} at time \ensuremath{t}. \\

\ensuremath{d_i(t)} &
Distance from vehicle \ensuremath{i} to the stop line at time
\ensuremath{t}. \\

\ensuremath{d_a(t)} &
Distance equivalent to: \ensuremath{d_a(t)=d_{\ell_a(t)}(t)}. \\

\ensuremath{\mathcal{M}} &
Set of maneuver intentions:
\ensuremath{\{\mathrm{left},\mathrm{straight},\mathrm{right}\}}. \\

\ensuremath{m_i} &
Declared maneuver intention of vehicle \ensuremath{i}. \\

\ensuremath{\Gamma_i} &
Crossing-zone footprint occupied by vehicle \ensuremath{i}. \\

\ensuremath{\chi_{ab}(t)} &
Binary geometric-conflict indicator between active approaches
\ensuremath{a} and \ensuremath{b} at time \ensuremath{t}. \\

\ensuremath{\mathcal{C}_t(a)} &
Conflict set of approach \ensuremath{a} at time \ensuremath{t}. \\

\ensuremath{R_a(t)} &
Role assigned to active approach \ensuremath{a} at time
\ensuremath{t}. \\

\ensuremath{v_i^{\mathrm{adv}}(t)} &
Advisory speed assigned to vehicle \ensuremath{i} at time
\ensuremath{t}. \\

\ensuremath{V_{\min}}, \ensuremath{V_{\max}} &
Minimum and maximum admissible advisory speed. \\

\ensuremath{a^*} &
Conflicting approach with right-of-way. \\

\ensuremath{\tau_{\mathrm{safe}}} &
Optional safety clearance buffer added to the estimated clearance time. \\

\ensuremath{\eta} &
Speed reduction factor used under \textsc{Share}. \\

\ensuremath{P_a} &
Queue pressures of approach \ensuremath{a}. \\

\ensuremath{v_\ell} &
Speed of the lane leader \ensuremath{\ell}. \\

\ensuremath{n_a} &
Number of queued vehicles on approach \ensuremath{a}. \\

\ensuremath{\bar{v}_a} &
Mean speed of vehicles on approach \ensuremath{a}. \\

\ensuremath{\alpha_{\mathrm{slow}}} &
Maximum slow-speed penalty used in the effective-delay calculation. \\

\ensuremath{\sigma} &
Conflict signature used as the lookup key. \\

\bottomrule
\end{tabular}
\end{table}

For each active approach $a \in \mathcal{A}_t$, let $\ell_a(t)$ denote
the leading vehicle in approach $a$ (i.e., the vehicle on approach $a$ closest to the
stop line). The distance of vehicle $i$ to the stop line is denoted by
$d_i(t)$, and the leader distance of approach $a$ is therefore
\begin{equation}
  d_a(t) = d_{\ell_a(t)}(t).
  \label{eq:leader_distance}
\end{equation}

Each vehicle $i$ declares a maneuver intention $m_i \in \mathcal{M}$,
where
\begin{equation}
  \mathcal{M} = \{\mathrm{left},\mathrm{straight},\mathrm{right}\}.
  \label{eq:movement_set}
\end{equation}
For a vehicle entering from approach
$a_i$, the pair $(a_i,m_i)$ determines the portion of the crossing zone
that the vehicle will occupy while executing its maneuver, denoted by:
\begin{equation}
  \Gamma_i = \Gamma(a_i,m_i).
  \label{eq:path_footprint}
\end{equation}
Two active approaches $a,b \in \mathcal{A}_t$ are in \textit{geometric conflict} if
the path footprints of their lane leaders overlap:

\begin{equation}
  \chi_{ab}(t) =
  \mathbf{1}\!\left[
    \Gamma_{\ell_a(t)} \cap \Gamma_{\ell_b(t)} \neq \emptyset
  \right],
  \qquad a \neq b.
  \label{eq:conflict_indicator}
\end{equation}
The conflict set of approach $a$ is then
\begin{equation}
  \mathcal{C}_t(a) =
  \left\{
    b \in \mathcal{A}_t \setminus \{a\}
    \;:\;
    \chi_{ab}(t)=1
  \right\}.
  \label{eq:conflict_set}
\end{equation}

The LLM arbitration layer assigns an approach-level role to each active
approach:
\begin{equation}
  R_a(t) \in
  \{\textsc{Clear},\textsc{Yield},\textsc{Share}\},
  \qquad a \in \mathcal{A}_t.
  \label{eq:approach_role}
\end{equation}
The role \textsc{Follow} is not issued by the LLM. It is assigned
deterministically by the execution layer to every non-leading vehicle on
an active approach after the lane leader's approach-level role has been
resolved.

\subsection{Vehicle Intent Representation}
\label{subsec:intent_representation}

In LIDSA, each vehicle declares its intent for two purposes:
it supplies the arbitration agent with the information needed
to resolve conflicts in a socially-aware manner, and it provides
the evaluation criterion against which arbitration quality is
measured at the end of each traversal.

A vehicle's intent captures four dimensions:

\begin{itemize}
  \item \textit{Spatial:} which route the vehicle intends to take through the intersection.
  \item \textit{Temporal:} whether the delay imposed on the vehicle remains within an acceptable budget relative to intersection load. This is the most consequential dimension, as delay is the primary cost of any intersection crossing.
  \item \textit{Priority:} the urgency class of the vehicle and its passenger occupancy. Priority class distinguishes vehicle types by
  social/legal weight, ranging from emergency vehicles such as ambulances, police cars, and fire trucks, through public transit and commercial
  operators, down to private cars. Occupancy scales the social cost of delay according to the number of passengers affected.
  \item \textit{Energy:} how sensitive the vehicle is to smooth, energy-efficient driving versus delay minimization, satisfied when the vehicle's speed profile through the intersection remains consistent with its declared driving preference.
\end{itemize}

Together, these four dimensions form a multi-objective criterion that reflects both the operational needs of the intersection and the individual preferences of each vehicle.

\subsection{Right-of-Way Assignments}
\label{subsec:row_assignments}

For each active approach, the LLM assigns one of three leader roles:
\textsc{Clear}, \textsc{Yield}, or \textsc{Share}. The fourth role,
\textsc{Follow}, is assigned deterministically by the execution layer to
non-leading vehicles. Table~\ref{tab:row_assignments} summarizes the
semantics of each role.

Assignments are produced using a fixed precedence order. First,
emergency vehicles trigger strict pre-emption: all geometrically
conflicting approaches yield, while non-conflicting approaches may
continue. Second, if two conflicting approaches are both saturated and
have comparable queue pressure, both receive \textsc{Share}. Third, all
remaining conflicts are resolved as \textsc{Clear}/\textsc{Yield} pairs
using priority class, weighted wait, and queue pressure as successive
tiebreakers. This ordering prevents congestion-balancing rules from
overriding emergency pre-emption.

\begin{table}[t]
\centering
\caption{Right-of-way roles in LIDSA. The LLM issues only
\textsc{Clear}, \textsc{Yield}, and \textsc{Share}; \textsc{Follow} is
assigned by the execution layer.}
\label{tab:row_assignments}
  \renewcommand{\arraystretch}{1.0}
\begin{tabular}{@{}lp{5.9cm}@{}}
\toprule
\textbf{Role} & \textbf{Meaning} \\
\midrule
\textsc{Clear}
  & The approach has priority and proceeds at free-flow speed unless
    limited by downstream constraints. Non-conflicting approaches are
    also assigned \textsc{Clear}. \\
\addlinespace[3pt]
\textsc{Yield}
  & The approach defers to a conflicting approach with higher priority.
    Its speed is chosen so that it reaches the stop line only after the
    winning approach has cleared the crossing zone. \\
\addlinespace[3pt]
\textsc{Share}
  & Two saturated conflicting approaches proceed concurrently at reduced,
    pressure-weighted speeds to prevent starvation. \\
\addlinespace[3pt]
\textsc{Follow}
  & A non-leading vehicle tracks the leader on the same approach with a
    safe headway buffer. This role is not produced by the LLM. \\
\bottomrule
\end{tabular}
\end{table}

\subsection{Right-of-Way-Roles-to-Speed Mapping}
\label{subsec:speed_mapping}

At each control step, the executor converts the assigned role of vehicle
$i$ into an advisory speed $v_i^{\mathrm{adv}}$ using the current
distance $d_i$ to the stop line and speed bounds $V_{\min}$ and
$V_{\max}$.

For \textsc{Clear}, the advisory speed is simply
\begin{equation}
  v_i^{\mathrm{adv}} = V_{\max}.
  \label{eq:speed_clear}
\end{equation}

For \textsc{Yield}, let $a^*$ denote the conflicting approach with
right-of-way. Its estimated clearance time is
\begin{equation}
  T_{a^*}^{\mathrm{clear}}(t)
  =
  \frac{d_{a^*}(t)}{v_{a^*}^{\mathrm{ref}}(t)}
  +
  \tau_{a^*}^{\mathrm{trav}}
  +
  \tau_{\mathrm{safe}},
  \label{eq:winner_clearance_time}
\end{equation}
where $v_{a^*}^{\mathrm{ref}}(t)=V_{\max}$ for a \textsc{Clear}
winner, $\tau_{a^*}^{\mathrm{trav}}$ is the traversal time through the
crossing zone, and $\tau_{\mathrm{safe}}$ is an optional clearance
buffer. In the reported experiments, $\tau_{\mathrm{safe}}=0$.

The yielding vehicle's raw advisory speed is
\begin{equation}
  \tilde{v}_i^{\mathrm{yield}}(t)
  =
  \frac{d_i(t)}{T_{a^*}^{\mathrm{clear}}(t)}.
  \label{eq:speed_yield_raw}
\end{equation}
The applied advisory is then
\begin{equation}
  v_i^{\mathrm{adv}}(t)
  =
  \begin{cases}
    0, &
    \tilde{v}_i^{\mathrm{yield}}(t) < V_{\min}, \\[3pt]
    \min\!\left(\tilde{v}_i^{\mathrm{yield}}(t), V_{\max}\right),
    & \text{otherwise}.
  \end{cases}
  \label{eq:speed_yield_clamp}
\end{equation}
Thus, speeds below $V_{\min}$ are interpreted as a full stop until the
next control update.

For \textsc{Share}, two conflicting approaches $a$ and $b$ proceed at
pressure-weighted reduced speeds:
\begin{equation}
  v_i^{\mathrm{adv}}
  =
  \eta V_{\max}
  \frac{P_a}{P_a + P_b},
  \qquad i \in a,
  \label{eq:speed_share}
\end{equation}
where $P_a$ and $P_b$ are the queue pressures defined in
Section~\ref{subsec:share}, and $\eta=0.85$ limits concurrent passage to
85\% of free-flow speed.

For \textsc{Follow}, vehicle $i$ matches the lane leader $\ell$ on the
same approach with a time headway buffer:
\begin{equation}
  v_i^{\mathrm{adv}}
  =
  \frac{d_i}{d_\ell / v_\ell + \delta_{\mathrm{gap}}},
  \label{eq:speed_follow}
\end{equation}
where $\delta_{\mathrm{gap}}=\SI{3.0}{\second}$.

\subsection{Priority Pre-emption}
\label{subsec:priority}

LIDSA enforces strict intent-priority pre-emption for emergency vehicles.
Let $a_e \in \mathcal{A}_t$ denote the active approach containing an
emergency vehicle, and let $b \in \mathcal{A}_t$ denote any active
approach whose role is being assigned. Recall that
$\mathcal{C}_t(a_e)$ is the set of active approaches whose planned
movements geometrically conflict with the emergency approach $a_e$.
Under emergency pre-emption, the role assigned to approach $b$ is

\begin{equation}
R_b(t) =
\begin{cases}
    \textsc{Clear}, & b = a_e,\\
    \textsc{Yield}, & b \in \mathcal{C}_t(a_e),\\
    \textsc{Clear}, & b \in \mathcal{A}_t \setminus
    \bigl(\mathcal{C}_t(a_e) \cup \{a_e\}\bigr).
\end{cases}
\label{eq:preemption}
\end{equation}

Thus, the emergency approach itself receives \textsc{Clear}; all
geometrically conflicting approaches receive \textsc{Yield}; and
non-conflicting approaches are allowed to continue with \textsc{Clear}.
For every approach assigned \textsc{Yield} under pre-emption, the yield
target is the emergency approach $a_e$.

This conflict-aware rule avoids blanket pre-emption: approaches that do
not geometrically conflict with the emergency vehicle are not required to
yield and therefore do not incur unnecessary delay. Transit vehicles are
also prioritized during conflict resolution, but they do not trigger
pre-emption. Instead, they follow the standard \textsc{Clear}/
\textsc{Yield} arbitration process, with priority class used as the
first tiebreaking criterion.

\subsection{Congestion Dissipation}
\label{subsec:share}

Under sustained demand on two conflicting approaches, repeated
\textsc{Yield} assignments can lead to starvation, with one approach
being deferred repeatedly while the other continues to clear.
\textsc{Share} addresses this case by allowing both approaches to move
at reduced speeds, with the assigned speeds weighted by their relative
queue pressure. For a conflicting pair $(a_i,a_j)$, \textsc{Share} is
activated when
\begin{equation}
  P_{a_i} > \theta_P \;\wedge\; P_{a_j} > \theta_P \;\wedge\;
  \lvert P_{a_i} - P_{a_j} \rvert < \Delta_P,
  \label{eq:share_condition}
\end{equation}
where $\theta_P$ is the saturation threshold and $\Delta_P$ is the
tolerance used to determine whether the two approaches have comparable
pressure.
The queue pressure on approach $a$ is defined as
\begin{equation}
  P_a = n_a \cdot \bar{w}_a^{\mathrm{eff}},
  \label{eq:pressure}
\end{equation}
where $n_a$ is the number of queued vehicles and
$\bar{w}_a^{\mathrm{eff}}$ is the effective delay on approach $a$:
\begin{equation}
\bar{w}_a^{\mathrm{eff}} =
w_a^{\mathrm{stop}} +
\alpha_{\mathrm{slow}}
\max\!\left(0,\;\frac{V_{\max}-\bar{v}_a}{V_{\max}}\right),  \label{eq:composite_delay}
\end{equation}
where $w_a^{\mathrm{stop}}$ is the mean stopped delay per vehicle on
approach $a$, $\bar{v}_a$ is the mean approach speed, and
$\alpha_{\mathrm{slow}}$ is the maximum slow-speed penalty. The
normalized speed-loss term increases as the approach speed decreases,
assigning no penalty when $\bar{v}_a = V_{\max}$ and a maximum penalty
of $\alpha_{\mathrm{slow}}$ when $\bar{v}_a = 0$. This allows crawling
vehicles that are not fully stopped to contribute to the effective
delay, preventing congestion under high-load conditions from being
underestimated by stopped-time delay alone.

\subsection{Conflict Signature and Memoized Arbitration Table}
\label{subsec:cache}

To avoid redundant LLM calls, LIDSA stores previously computed
right-of-way assignments in a Memoized Arbitration Table (MAT). Because
the executor recomputes advisory speeds from live distances at every
control step, the MAT stores only symbolic role assignments rather than
speeds.

The MAT is keyed by a discrete conflict signature $\sigma$:

\begin{equation}
  \sigma = \left\langle\,
    \left\langle a,\; \mathrm{pc}_a^{\mathrm{dom}},\; r_a,\;
    b_a^{P},\; b_a^{E},\; b_a^{Q},\; b_a^{W},\; b^{\mathrm{net}}
    \right\rangle_{a \in \mathcal{A}},\;
    a^*
  \,\right\rangle,
  \label{eq:signature}
\end{equation}
where $\mathrm{pc}_a^{\mathrm{dom}}$ is the dominant priority class
on approach $a$; $r_a \in \{0,1,2\}$ is a class-normalized urgency
band defined as:
\begin{equation}
  r_a = \left\lfloor \frac{w_a}{\delta_T^{\mathrm{class}}} \right\rfloor,
  \label{eq:urgency_band}
\end{equation}
with $\delta_T^{\mathrm{class}}$ the class-specific delay budget;
$b_a^{P}$, $b_a^{E}$, and $b_a^{Q}$ are the pressure, energy
preference, and queue depth bands, respectively; and the absolute
wait band:
\begin{equation}
b_a^{W} =
\min\!\left(
\left\lfloor \frac{w_a}{\Delta_W} \right\rfloor,
B_W^{\max}
\right),
\label{eq:wait_band}
\end{equation}
which discretizes the waiting time into $\Delta_W$ (5\,s intervals) and caps the resulting band index at $B_W^{\max}=24$, corresponding to waits of 120\,s or longer.
The sequence $\sigma$ is then encoded as an ordered list of discrete
tokens and used as the lookup key for the MAT.

The current right-of-way holder $a^*$ is appended to $\sigma$ to break
directional symmetry between mirror-image states.

\subsection{Safety Watchdog}
\label{subsec:watchdog}

A tile-based collision monitor runs at every control step as a
deterministic safety backstop, independent of the LLM and MAT. If two
vehicles with geometrically conflicting movements occupy the same tile,
the later-arriving vehicle is overridden to a hold speed of
$V_{\mathrm{hold}}=\SI{2.0}{m/s}$ until the tile clears. This prevents
collisions even if an LLM assignment is incorrect or a cached MAT entry
is stale.

\begin{table}[!t]
  \caption{Benchmark Scenario Groups}
  \label{tab:scenario_groups}
  \centering
  \footnotesize
  \renewcommand{\arraystretch}{1.0}
  \setlength{\tabcolsep}{10pt}
  \begin{threeparttable}
    \begin{tabular}{@{}clcc@{}}
      \toprule
      \textbf{Group} & \textbf{Focus} & \textbf{IDs} & \textbf{Runs} \\
      \midrule
      A & Correctness under clear rules     & 1--6   & 3    \\
      B & Tie-breaking (wait, pressure)     & 7--10  & 3    \\
      C & \textsc{Share} role \& congestion & 11--14 & 3    \\
      D & Output format \& reliability      & 15--18 & 3    \\
      E & Operational fitness / determinism & 19--20 & 3/10\tnote{a} \\
      \midrule
        & \textbf{Total calls / model}      &        & \textbf{67} \\
      \bottomrule
    \end{tabular}
    \begin{tablenotes}[flushleft]
      \scriptsize
      \item[a] Scenario~19 executed 3~runs; Scenario~20
               (output determinism) executed 10~runs.
    \end{tablenotes}
  \end{threeparttable}
\end{table}

\section{LLM Benchmark for Right-of-Way Arbitration}
\label{sec:llm_benchmark}

\subsection{Motivation}
\label{subsec:benchmark_motivation}

The right-of-way arbitration module requires an LLM to assign one of three approach-level roles, \textsc{Clear}, \textsc{Yield}, or \textsc{Share}. This constrained task has a fixed output space, deterministic rules, structured JSON requirements, and direct effects on vehicle speed advisories. General benchmarks such as MMLU~\cite{hendrycks2021measuring} and HumanEval~\cite{chen2021evaluatinglargelanguagemodels} do not capture these requirements, while AgentDrive~\cite{ferrag2026agentdrive} is broader than the single-intersection setting considered here.

Model suitability is therefore evaluated using three criteria: logical determinism, i.e., consistent assignments for structurally identical conflict states; structured-output reliability, i.e., parseable JSON without fallback handling; and real-time latency, since delays beyond \SI{5}{\second} can make advisories stale despite asynchronous execution. Because no existing benchmark jointly evaluates these properties under intersection-control conditions, we use a purpose-built benchmark with 20 scenarios across five difficulty groups (Table~\ref{tab:scenario_groups}). Each model is evaluated over 67 API calls across three independent runs, including ten repeated calls for the output-determinism scenario.

\subsection{LLM Model Selection and Scoring}
\label{subsec:model_selection}

We evaluated five API-accessible models: Gemini 2.5 Flash Lite, Gemini 2.5 Flash, GPT-4o, Claude Haiku 4.5, and Kimi K2. These models span hybrid reasoning with switchable chain-of-thought, dense multimodal reasoning, compact instruction tuning, and open-weight Mixture-of-Experts reasoning. All models were queried at temperature zero with structured JSON output and a maximum of 8\,192 output tokens. Kimi K2 used a \SI{60}{\second} timeout to account for MoE routing overhead, while all other models used \SI{30}{\second}.

Each model receives a composite score $S$ based on logic accuracy, JSON parse rate, latency, and role-safety rate:

\begin{equation}
S =
w_{\ell}S_{\ell}
+ w_{j}S_{j}
+ w_{\lambda}L(\bar{\lambda})
+ w_{s}S_{s},
\label{eq:composite}
\end{equation}

where $S_{\ell}$, $S_{j}$, and $S_{s}$ denote logic accuracy, JSON parse rate, and role-safety rate, respectively. The latency term penalizes mean response time $\bar{\lambda}$ relative to a zero-score threshold $\lambda_{0}$:

\begin{equation}
L(\bar{\lambda}) =
\max\left(0,\;1-\frac{\bar{\lambda}}{\lambda_{0}}\right).
\label{eq:latency_score}
\end{equation}

Weights are set to $w_{\lambda}=0.40$, $w_{\ell}=0.30$, $w_{s}=0.20$, and $w_{j}=0.10$, prioritizing real-time responsiveness and logical correctness over output-format reliability. A model is disqualified if $S_{\ell}<0.50$ regardless of its composite score.

The latency threshold is set to $\lambda_{0}=\SI{5000}{\milli\second}$. This value reflects the intersection geometry and arbitration cadence: vehicles enter the scheduling horizon 400\,m from the stop line and, at 13.89\,m/s or 50\,km/h, reach it in approximately 28\,s. Since arbitration calls are issued every \SIrange{20}{30}{\second}, responses exceeding 5\,s may leave insufficient time for speed advisories to influence approach behavior before vehicles enter the near-zone conflict region.

\subsection{Benchmark Results}
\label{subsec:results}

Table~\ref{tab:results-scores} reports the benchmark results. The top four models form a high-correctness tier, each achieving logic accuracy above 98.8\% with perfect JSON parse and role-safety rates. Their ranking is therefore determined mainly by latency. Kimi K2 forms a lower-performing tier, with mean latency exceeding the 5\,s threshold and reduced JSON parse and role-safety rates.

Gemini 2.5 Flash Lite achieved the highest composite score, 93.3\%, due to its lowest mean latency, \SI{1592}{\milli\second}, zero calls above 5\,s, and near-perfect logic accuracy of 99.3\%. It is therefore selected as the primary LLM arbitrator for the SUMO evaluation.

\begin{table*}[!t]
  \renewcommand{\arraystretch}{1.0}
  \caption{LLM Benchmark Results}
  \label{tab:results-scores}
  \centering
  \footnotesize
  \setlength{\tabcolsep}{9pt}
  \begin{threeparttable}
    \begin{tabular}{@{}lrrrrr@{}}
      \toprule
      \textbf{Metric}
        & \textbf{Gemini 2.5 Flash Lite}
        & \textbf{Gemini 2.5 Flash}
        & \textbf{GPT-4o}
        & \textbf{Claude Haiku 4.5}
        & \textbf{Kimi K2} \\
      \midrule

      \multicolumn{6}{@{}l}{\textit{Score and reliability:}} \\
      Composite score (\%) & \textcolor{red}{93.3} & 92.1 & 88.9 & 87.3 & 75.7 \\
      Logic accuracy (\%)  & 99.3 & \textcolor{red}{99.8} & 99.0 & 98.8 & 91.9 \\
      JSON parse rate (\%) & \textcolor{red}{100.0} & \textcolor{red}{100.0} & \textcolor{red}{100.0} & \textcolor{red}{100.0} & 95.0 \\
      Role-safety rate (\%) & \textcolor{red}{100.0} & \textcolor{red}{100.0} & \textcolor{red}{100.0} & \textcolor{red}{100.0} & 95.0 \\

      \multicolumn{6}{@{}l}{\textit{Latency and efficiency:}} \\
      Mean latency (ms) & \textcolor{red}{1{,}592} & 1{,}910 & 2{,}633 & 2{,}983 & 5{,}175 \\
      P95 latency (ms)  & 2{,}613 & \textcolor{red}{2{,}542} & 4{,}243 & 4{,}523 & 6{,}898 \\
      Calls \textgreater 5{,}000\,ms & \textcolor{red}{0} & 2 & 1 & 2 & 22 \\
      Total tokens / call & 3{,}331 & 3{,}344 & 3{,}043 & 4{,}258 & \textcolor{red}{2{,}874} \\
      Throughput (tok/s) & \textcolor{red}{146.9} & 126.7 & 70.9 & 85.8 & 24.7 \\

      \bottomrule
    \end{tabular}

    \begin{tablenotes}[flushleft]
      \scriptsize
      \item Scoring weights: latency 40\%, logic accuracy 30\%, role safety 20\%, and JSON parse rate 10\%.
      The latency score decreases linearly from 100\% at 0\,ms to 0\% at 5{,}000\,ms.
      \item Red indicates the best value in each row. Each model was evaluated over n=67 API calls.
      \item Reasoning tokens were zero for all models because thinking was disabled or not applicable.
    \end{tablenotes}
  \end{threeparttable}
\end{table*}

\section{Experiment Setup}
\label{sec:setup}

This section describes the simulation environment, network
topology, traffic demand scenarios, baseline controllers,
and evaluation metrics used to assess LIDSA. All experiments
are implemented in Python using the SUMO microscopic traffic
simulator~\cite{sumo2018} interfaced via the TraCI API.

\begin{table}[ht]
  \centering
  \caption{Simulation Parameters}
  \label{tab:sim_params}
  \renewcommand{\arraystretch}{1.0}
  \setlength{\tabcolsep}{2pt}
  \begin{tabular}{p{4cm}l}
    \toprule
    \textbf{Parameter} & \textbf{Value} \\
    \midrule
    \textit{Environment:} & \\
    Simulator          & SUMO \\
    Step length        & \SI{1}{\second} \\
    Evaluation window  & \SI{3600}{\second} \\
    Random seeds       & \ensuremath{\{7,\; 41,\; 129\}} \\
    \midrule
    \textit{Network:} & \\
    Intersection type  & Isolated four-way \\
    Edges              & 8 (4 inbound, 4 outbound) \\
    Edge length        & \SI{600}{\metre} \\
    Speed limit        & \SI{13.89}{\mps} \\
    Conflict zone      & \SI{12}{\metre} \ensuremath{\times} \SI{12}{\metre} \\
    Advisory horizon   & \SI{400}{\metre} \\
    Near-zone radius   & \SI{200}{\metre} \\
    \ensuremath{V_{\min}} & \SI{3.0}{\mps} \\
    \ensuremath{V_{\max}} & \SI{13.89}{\mps} \\
    \midrule
    \textit{LIDSA:} & \\
    LLM temperature    & \ensuremath{T = 0.0} \\
    LLM seed           & 42 \\
    Max output tokens  & 2048 \\
    Query cadence      & \SI{30}{\second} (30 steps) \\
    \bottomrule
  \end{tabular}
\end{table}

\subsection{Simulation Environment}
\label{subsec:sim_env}

Experiments are conducted in SUMO with a \SI{1}{\second} step length
over a \SI{3600}{\second} evaluation window. The network is a
synthetic isolated four-way intersection, comprising four inbound edges and four
outbound edges, each \SI{600}{\metre} long with a single lane and a speed limit of \SI{13.89}{\mps} (\SI{50}{\kmph}). The conflict zone is modeled as a
\SI{12}{\metre}\,$\times$\,\SI{12}{\metre} box centered at the origin.

Each scenario configuration is evaluated using three independent random
seeds ($\{7, 41, 129\}$), and all reported metrics are averaged across
these runs.

The simulation includes three vehicle classes with distinct kinematic
and priority profiles: \textit{emergency}, \textit{transit}, and
\textit{normal} vehicles. Each vehicle is assigned an energy-preference parameter,
\ensuremath{\alpha \sim \mathcal{U}(0,1)}, which modulates the
deceleration comfort threshold used in the energy intent sub-metric
described in Section~\ref{subsec:metrics_intent}. Emergency vehicles,
including ambulances and police vehicles, are spawned at a fixed rate
of 20~veh/h.

\begin{table}[!b]
  \centering
  \caption{Traffic Demand Scenarios}
  \label{tab:scenarios}
  \setlength{\tabcolsep}{4pt}
  \begin{tabular}{lcccc}
    \toprule
    \textbf{Scenario} & \textbf{NS Straight} & \textbf{EW Straight} &
    \textbf{Turns} & \textbf{\ensuremath{v/c}} \\
    & (veh/h) & (veh/h) & (veh/h) & \\
    \midrule
    Low    & 150 & 120 &  40 & 0.24 \\
    Medium & 400 & 300 & 100 & 0.63 \\
    High   & 600 & 500 & 150 & 0.94 \\
    \bottomrule
  \end{tabular}
\end{table}

\subsection{Traffic Load Scenarios}
\label{subsec:scenarios}

Three traffic-load scenarios are defined to represent operating
conditions ranging from undersaturated to near-saturated flow, as
summarized in Table~\ref{tab:scenarios}. Load is specified in vehicles
per hour for each movement, with \textit{Straight-through} and \textit{turning} flows applied
symmetrically across the four approaches.

The volume-to-capacity ratio, $v/c$, is computed as

\begin{equation}
v/c = \max\!\left(
    \frac{q_{NS}}{c_{NS}},\;
    \frac{q_{EW}}{c_{EW}}
\right).
\end{equation}

Here, $q_{NS}$ and $q_{EW}$ denote the total directional loads
in veh/h, while $c_{NS}$ and $c_{EW}$ denote the corresponding
single-lane directional capacities:

\begin{equation}
c_{NS} = \frac{g_{NS}}{C} \cdot s \cdot N,
\qquad
c_{EW} = \frac{g_{EW}}{C} \cdot s \cdot N.
\end{equation}

In this calculation, $g_{NS} = g_{EW} = \SI{30}{\second}$ is the
effective green time, $C = \SI{68}{\second}$ is the cycle length,
$s = \SI{1800}{veh\per\hour\per lane}$ is the saturation flow rate,
and $N = 1$ is the number of lanes per approach.

\subsection{Baseline Controllers}
\label{subsec:baselines}

LIDSA is evaluated against four baseline controllers: \textit{Fixed-cycle}, \textit{Adaptive Signal}, \textit{Autonomous Intersection Management} (AIM), and \textit{Green Light Optimal Speed Advisory} (GLOSA); each representing different and varied approaches:

\textit{Fixed-Cycle (\textit{Fixed}):} A pre-timed two-phase signal
controller with symmetric green splits: $g_{NS} = g_{EW} = 30$\,s,
yellow $y = 4$\,s, and cycle $C = 68$\,s. No adaptation to
real-time demand.

\textit{SCATS}: A three-plan adaptive
controller that selects among pre-defined timing plans based on the
degree of saturation $d_s$ ($d_s=v/c$).
Plans are indexed by $d_s$ thresholds: Plan~1 ($d_s < 0.50$,
$g_{NS} = 20$\,s, $g_{EW} = 20$\,s), Plan~2 ($d_s < 0.80$,
$g_{NS} = 35$\,s, $g_{EW} = 25$\,s), and Plan~3
($d_s \geq 0.80$, $g_{NS} = 50$\,s, $g_{EW} = 40$\,s).

\textit{AIM}: A
signal-free reservation-based controller operating on a
$10 \times 10$ tile grid spanning a
$\SI{30}{\metre} \times \SI{30}{\metre}$ conflict zone at
$\SI{3}{\metre}$ tile resolution. Vehicles within \SI{60}{\metre}
of the intersection box submit reservation requests; the server
finds the earliest conflict-free time slot via linear scan and
commits a grant. Reservations are pruned every 10 steps. Vehicles
beyond \SI{60}{\metre} travel at $V_{max}$.

\textit{GLOSA}: A
rule-based speed advisory controller that uses a virtual fixed-cycle
SPaT reference as though the intersection were signal-controlled.
The advisory speed for vehicle $i$ at distance $d_i$ from the stop
line is:
\begin{equation}
v_{adv,i} =
\begin{cases}
  v_{max} & \text{if currently in green window} \\[4pt]
  \displaystyle\frac{d_i}{t_{green,i}} & \text{otherwise}
\end{cases}    
\end{equation}
where $t_{green,i}$ is the time until the next green window onset
for approach $i$, derived from the virtual cycle position
$t_{cycle} = \text{step} \bmod C$. Conflict resolution is FIFO by
distance: when two vehicles from conflicting approaches would arrive
within $\Delta t_{min} = \SI{3}{\second}$ of each other, the nearer
vehicle proceeds and the farther vehicle targets the subsequent green
window. The intersection operates permanently green; the virtual
SPaT is advisory only.

\begin{table}[!b]
  \centering
  \caption{Vehicle Type Profiles}
  \label{tab:vtypes}
  \setlength{\tabcolsep}{4pt}
  \begin{tabular}{lcccclr}
    \toprule
    \textbf{Type} & \textbf{Length} & \textbf{Accel} &
    \textbf{Decel} & \textbf{$v_{\max}$} &
    \textbf{Priority} & \textbf{Occ.} \\
    & (m) & (m/s$^2$) & (m/s$^2$) & (m/s) & Class & \\
    \midrule
    Car       & 5  & 2.6 & 4.5 & 13.89 & NORMAL    & 1  \\
    Bus       & 12 & 1.2 & 3.5 & 11.11 & TRANSIT   & 35 \\
    Ambulance & 6  & 3.0 & 5.0 & 16.67 & EMERGENCY & 2  \\
    \bottomrule
  \end{tabular}
\end{table}
\subsection{LIDSA Controller Configuration}
\label{subsec:LIDSA_config}

LIDSA operates on a permanently green intersection with no physical
signal. The LLM backend is configured with temperature $T = 0.0$,
seed 42, and a maximum of 2048 output tokens. The query cadence is
$\Delta t = 30$ steps ($\SI{30}{\second}$).

The advisory horizon extends \SI{400}{\metre} upstream of the stop line;
vehicles outside this range continue at the unmodified maximum speed,
$V_{max}$. Right-of-way assignments are applied within the near zone,
defined as the \SI{200}{\metre} region upstream of the stop line. Vehicle
speeds are constrained to the interval $[V_{min},\,V_{max}] =
[\SI{3.0}{m/s},\,\SI{13.89}{m/s}]$.

\subsection{Evaluation Metrics}
\label{subsec:metrics}

Metrics are computed for each controller--scenario--seed combination
and averaged over three random seeds. We report four metric families:
traffic efficiency, queue suppression, intent satisfaction, and
energy/emissions.

\textit{Traffic efficiency} is measured using throughput, mean control
delay, mean waiting time, and mean speed. Throughput is the number of
vehicles completing their trip within the evaluation window. Mean
control delay is the primary HCM-based efficiency measure and is
computed over completed trips as
\begin{equation}
    \bar{d} =
    \frac{1}{N_{\mathrm{arr}}}
    \sum_{i=1}^{N_{\mathrm{arr}}}
    \max\!\left(0,\; t^{\mathrm{travel}}_i -
    t^{\mathrm{freeflow}}_i\right),
    \label{eq:control_delay}
\end{equation}
where $t^{\mathrm{travel}}_i=t^{\mathrm{arr}}_i-t^{\mathrm{dep}}_i$ and
$t^{\mathrm{freeflow}}_i=\ell_i/V_{\max}$. Mean waiting time is the
time-averaged waiting duration of active vehicles whose speed falls
below \SI{0.1}{\metre\per\second}, sampled every 10 simulation steps.
Mean speed is the corresponding time-averaged network speed over active
vehicles.

\textit{Queue suppression} is evaluated using average and peak queue
lengths. Let $h_{e,k}$ denote the number of halting vehicles on inbound
edge $e$ at sample $k$, and let $\mathcal{E}_{\mathrm{in}}$ be the set
of inbound edges. The average and peak queue lengths are
\begin{equation}
\bar{q} =
\frac{1}{N_{\mathrm{samp}}}
\sum_{k=1}^{N_{\mathrm{samp}}}
\sum_{e \in \mathcal{E}_{\mathrm{in}}} h_{e,k},
\qquad
q_{\max} =
\max_k
\sum_{e \in \mathcal{E}_{\mathrm{in}}} h_{e,k}.
\label{eq:queue_metrics}
\end{equation}

\textit{Intent satisfaction} measures whether each completed trip
satisfies four declared objectives: spatial route fidelity, temporal
delay compliance, priority service, and energy-aware smoothness. The
vehicle-level score is
\begin{equation}
I_i =
0.20\,S^{\mathrm{sp}}_i
+ 0.40\,S^{\mathrm{tm}}_i
+ 0.20\,S^{\mathrm{pr}}_i
+ 0.20\,S^{\mathrm{en}}_i,
\label{eq:intent_vehicle}
\end{equation}
and the fleet-level intent satisfaction rate is
\begin{equation}
\bar{I} =
\frac{1}{|\mathcal{V}_{\mathrm{arr}}|}
\sum_{i \in \mathcal{V}_{\mathrm{arr}}} I_i.
\label{eq:intent_fleet}
\end{equation}
The spatial score verifies that the vehicle exits through the expected
outbound edge. The temporal score checks whether control delay remains
within a demand-dependent budget of 30, 60, or 120\,s for low, moderate,
and high volume-to-capacity conditions, respectively. The priority score
applies to emergency and transit vehicles and checks whether near-zone
delay remains below the corresponding class threshold. The energy score
checks whether deceleration and stop-frequency limits are satisfied for
vehicles with nonzero energy preference.

\textit{Energy and emissions} are measured using average fuel
consumption, kinetic energy loss, and average stops per vehicle. Fuel
consumption is estimated using a Vehicle Specific Power (VSP) bin model.
For vehicle $i$ at step $k$,
\begin{equation}
VSP_{i,k} =
v_{i,k}\!\left(1.1\,a_{i,k}+0.132\right)
+ 3.02 \times 10^{-4} v_{i,k}^{3},
\label{eq:vsp}
\end{equation}
where $v_{i,k}$ and $a_{i,k}$ are instantaneous speed and acceleration.
The instantaneous fuel rate $\dot{m}_{i,k}$ is assigned from the
eight-bin VSP fuel model, and average fuel consumption is
\begin{equation}
\bar{F} =
\frac{1}{|\mathcal{V}_{\mathrm{arr}}|}
\sum_{i \in \mathcal{V}_{\mathrm{arr}}}
\sum_k \dot{m}_{i,k}\Delta t,
\qquad \Delta t=\SI{1}{\second}.
\label{eq:fuel}
\end{equation}
Kinetic energy loss is computed over deceleration events as
\begin{equation}
\bar{E}_{\mathrm{KE}} =
\frac{1}{|\mathcal{V}_{\mathrm{arr}}|}
\sum_{i \in \mathcal{V}_{\mathrm{arr}}}
\sum_{k:\, v_{i,k}<v_{i,k-1}}
\frac{1}{2}m
\left(v_{i,k-1}^{2}-v_{i,k}^{2}\right),
\label{eq:ke_loss}
\end{equation}
with $m=\SI{1500}{\kilo\gram}$. Average stops per vehicle is reported
as a diagnostic smoothness metric and counts moving-to-stopped
transitions, where stopped is defined as
$v<\SI{0.1}{\metre\per\second}$.

\subsection{Experimental Protocol}
\label{subsec:protocol}

Each of the five controllers (Fixed, SCATS, AIM, GLOSA, LIDSA) is evaluated under each of the three traffic
scenarios (Low, Medium, High) across three random seeds,
yielding $5 \times 3 \times 3 = 45$ simulation runs in
total. Each run is \SI{3600}{\second} of simulated time.
Metrics are sampled every 10 simulation steps. Scalar KPIs
are averaged across seeds; standard deviations are computed
but not reported in the main tables due to low seed-to-seed
variance (all $\sigma < 5\%$ of the mean for primary metrics).
A checkpoint manager persists completed run results to disk,
enabling incremental execution across sessions.

For LLM-based controllers, the model API is queried
asynchronously in a background thread to avoid blocking the
TraCI simulation loop. If a query does not complete within
the $\SI{30}{\second}$ timeout, a CLEAR fallback is applied
to all near-zone approaches. The policy cache is initialized
empty at the start of each run and accumulates role decisions for the duration of that run.

\begin{table*}[ht]
  \centering
  \begin{threeparttable}
  \caption{
  Performance summary across load scenarios. LIDSA results are simulated using the \texttt{gemini2.5-flash-lite} model. Best-performing values are highlighted in red.
  }
  \label{tab:results_summary}
  \renewcommand{\arraystretch}{1.0}
  \setlength{\tabcolsep}{3.6pt}
  \footnotesize
  \begin{tabular}{lrrrrrrrrrrrrrrr}
    \toprule
    & \multicolumn{5}{c}{\textbf{Low} (\ensuremath{v/c = 0.24})}
    & \multicolumn{5}{c}{\textbf{Medium} (\ensuremath{v/c = 0.63})}
    & \multicolumn{5}{c}{\textbf{High} (\ensuremath{v/c = 0.94})} \\
    \cmidrule(lr){2-6}\cmidrule(lr){7-11}\cmidrule(lr){12-16}
    \textbf{Metric}
      & \textbf{Fixed} & \textbf{SCATS} & \textbf{AIM} & \textbf{GLOSA} & \textbf{LIDSA}
      & \textbf{Fixed} & \textbf{SCATS} & \textbf{AIM} & \textbf{GLOSA} & \textbf{LIDSA}
      & \textbf{Fixed} & \textbf{SCATS} & \textbf{AIM} & \textbf{GLOSA} & \textbf{LIDSA} \\
    \midrule
    \multicolumn{16}{l}{\textit{Traffic Efficiency}} \\
    Throughput (veh)
      & 717 & 716 & 717 & \textcolor{red}{717} & \textcolor{red}{717}
      & 1417 & 1382 & 1733 & 1733 & \textcolor{red}{1786}
      & 1401 & 1431 & 1707 & \textcolor{red}{1806} & 1794 \\
    Ctrl.\ Delay (s/veh)
      & 25.1 & 24.1 & 15.1 & 23.8 & \textcolor{red}{10.5}
      & 203.7 & 199.2 & 87.2 & 72.9 & \textcolor{red}{22.2}
      & 360.4 & 344.7 & 287.0 & 264.1 & \textcolor{red}{261.7} \\
    \ensuremath{\Delta} vs.\ Fixed (\%)\tnote{1}
      & --- & 3.9 & 39.6 & 4.9 & \textcolor{red}{57.9}
      & --- & 2.2 & 57.2 & 64.2 & \textcolor{red}{89.1}
      & --- & 4.3 & 20.4 & 26.7 & \textcolor{red}{27.4} \\
    Mean Speed (km/h)
      & 38.9 & 39.3 & 43.0 & 39.4 & \textcolor{red}{44.7}
      & 15.5 & 15.5 & 25.5 & 28.3 & \textcolor{red}{40.6}
      & 10.3 & 10.6 & 12.3 & 13.2 & \textcolor{red}{13.3} \\
    LOS\tnote{2}
      & C & C & B & C & \textcolor{red}{B}
      & F & F & F & F & \textcolor{red}{C}
      & F & F & F & F & \textcolor{red}{F} \\
    \midrule
    \multicolumn{16}{l}{\textit{Queue Management}} \\
    Avg Queue (veh)
      & 1.92 & 1.82 & \textcolor{red}{0.00} & \textcolor{red}{0.00} & 0.01
      & 38.1 & 52.9 & 1.09 & 4.22 & \textcolor{red}{0.62}
      & 68.9 & 79.0 & 11.0 & \textcolor{red}{8.90} & 9.92 \\
    Peak Queue (veh)
      & 6.3 & 6.3 & \textcolor{red}{0.0} & \textcolor{red}{0.0} & 1.0
      & 62.0 & 84.7 & 15.3 & 23.0 & \textcolor{red}{7.0}
      & 94.0 & 104.7 & 93.0 & 41.3 & \textcolor{red}{37.0} \\
    Avg Wait (s/veh)
      & 0.99 & 1.01 & \textcolor{red}{0.00} & \textcolor{red}{0.00} & \textcolor{red}{0.00}
      & 3.89 & 11.1 & 0.09 & 0.37 & \textcolor{red}{0.08}
      & 4.75 & 9.70 & 3.90 & \textcolor{red}{0.31} & 0.34 \\
    \midrule
    \multicolumn{16}{l}{\textit{Intent Satisfaction}} \\
    Overall (\%)
      & 69.1 & 67.5 & 87.5 & 76.1 & \textcolor{red}{89.5}
      & 52.3 & 60.3 & 56.1 & 61.2 & \textcolor{red}{86.2}
      & 30.8 & \textcolor{red}{40.0} & 33.7 & 35.8 & 37.1 \\
    Temporal (\%)
      & 62.0 & 65.8 & 92.4 & 65.6 & \textcolor{red}{96.8}
      & 48.9 & 61.3 & 44.0 & 55.1 & \textcolor{red}{95.0}
      & 4.17 & \textcolor{red}{29.1} & 4.20 & 6.57 & 7.70 \\
    Priority (\%)
      & 31.4 & 15.7 & \textcolor{red}{56.8} & \textcolor{red}{56.8} & \textcolor{red}{56.8}
      & 0.00 & 15.7 & 5.90 & 7.97 & \textcolor{red}{51.0}
      & 0.00 & 2.57 & 5.80 & 8.37 & \textcolor{red}{8.30} \\
    Energy (\%)
      & 90.3 & 90.0 & 95.8 & 92.2 & \textcolor{red}{96.9}
      & 64.0 & 63.1 & 86.4 & 88.0 & \textcolor{red}{90.0}
      & 45.8 & 39.2 & 54.1 & 57.6 & \textcolor{red}{62.0} \\
    \midrule
    \multicolumn{16}{l}{\textit{Energy and Emissions}} \\
    Fuel (g/veh)
      & 89.6 & 88.6 & 80.9 & 85.9 & \textcolor{red}{66.9}
      & 169.6 & 173.9 & 120.7 & 116.3 & \textcolor{red}{86.9}
      & 229.6 & 230.2 & 174.7 & \textcolor{red}{173.6} & 175.0 \\
    KE Loss (kJ/veh)
      & 119.4 & 114.7 & 96.1 & 99.9 & \textcolor{red}{58.4}
      & 136.7 & 135.7 & 149.3 & 150.0 & \textcolor{red}{108.9}
      & 152.6 & 145.9 & \textcolor{red}{127.8} & 139.8 & 140.9 \\
    Stops/veh
      & 1.0 & 1.0 & 1.0 & \textcolor{red}{0.0} & 1.0
      & 4.39 & 4.00 & 1.15 & 1.43 & \textcolor{red}{1.06}
      & 6.30 & 4.74 & \textcolor{red}{1.65} & 1.84 & 2.06 \\
    \bottomrule
  \end{tabular}
  \begin{tablenotes}[flushleft]
    \footnotesize
    \item[1] \ensuremath{\Delta} denotes the percentage reduction in mean control delay relative to the Fixed-Cycle controller.
    \item[2] LOS grades follow Highway Capacity Manual (HCM) control-delay thresholds~\cite{hcm2022}.
  \end{tablenotes}
  \end{threeparttable}
\end{table*}

\section{Performance Analysis}
\label{sec:performance}

We evaluate LIDSA against four baselines across three demand levels:
Low, Medium, and High, corresponding to \ensuremath{v/c=0.24},
\ensuremath{v/c=0.63}, and \ensuremath{v/c=0.94}, respectively.
The baselines are a fixed-cycle signal controller (Fixed), an adaptive
signal controller (SCATS), an autonomous intersection manager (AIM),
and a rule-based speed advisory controller (GLOSA). All results are
averaged over three independent random seeds
\ensuremath{\{7,\,41,\,129\}}. Table~\ref{tab:results_summary}
summarizes the main results; the following subsections discuss them by
metric family.

\subsection{Traffic Efficiency}
\label{subsec:traffic_efficiency}

\begin{figure}[!t]
  \centering
  \includegraphics[width=\columnwidth]{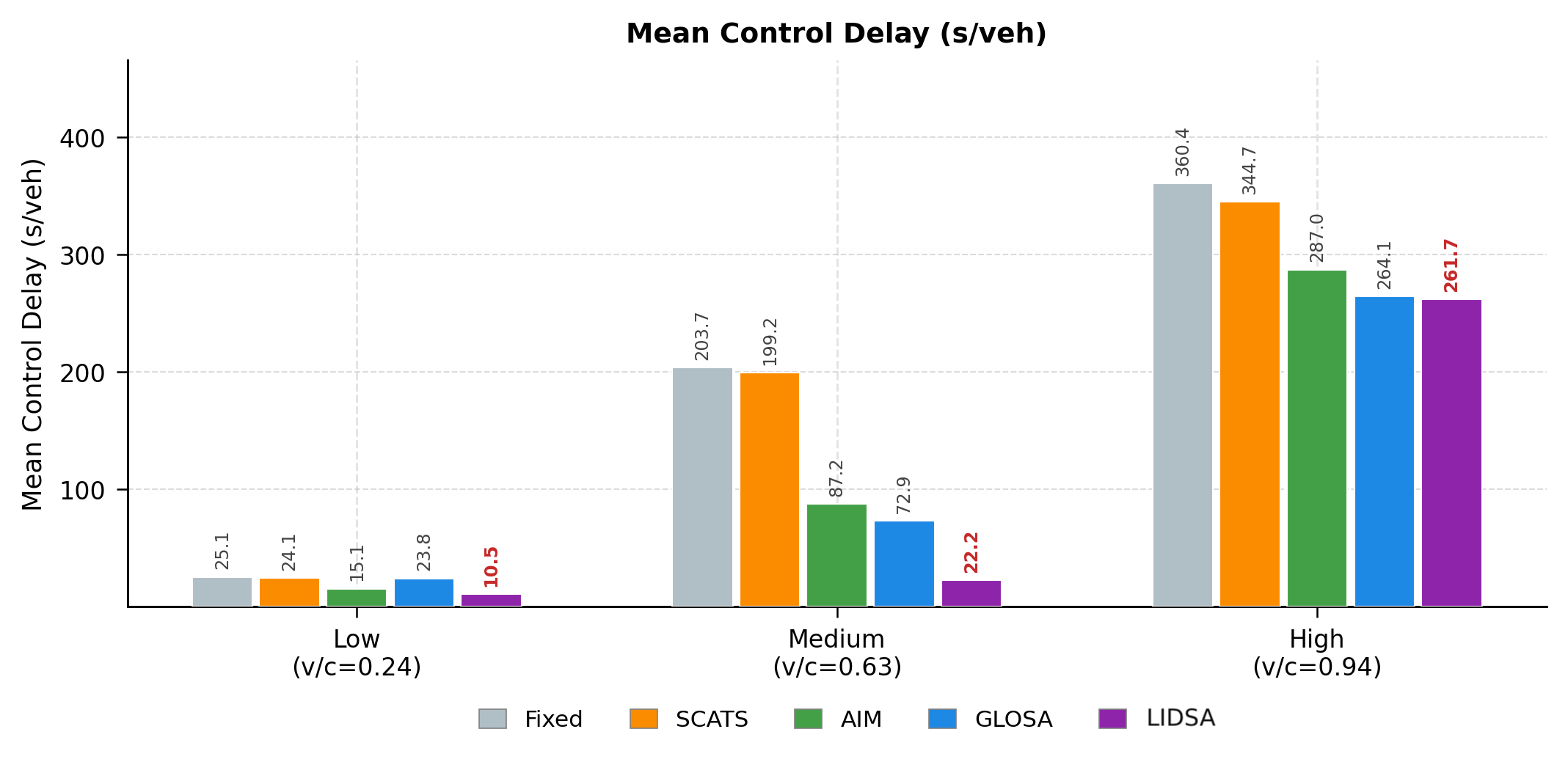}
  \caption{Mean control delay (s/veh). }
  \label{fig:delay}
\end{figure}

Traffic-efficiency results show that LIDSA maintains high throughput
while substantially reducing delay, waiting time, and speed degradation.
At Low load, throughput is comparable across all controllers
at approximately 717 vehicles, indicating that the network is not
supply-constrained. At Medium load, LIDSA achieves the highest throughput
with 1786 vehicles. At High load, throughput converges among the best
signal-free and advisory-based methods, with GLOSA slightly ahead at
1806 vehicles and LIDSA close behind at 1794 vehicles. Fixed and SCATS
degrade between Medium and High load because signalized operation
becomes oversaturated.

Mean control delay is the clearest efficiency differentiator. As shown
in Fig.~\ref{fig:delay}, LIDSA achieves the lowest delay at every load
level: 10.5 s/veh at Low load, 22.2 s/veh at Medium load, and
261.7 s/veh at High load. Relative to Fixed, these correspond to
reductions of 58.2\%, 89.1\%, and 27.4\%, respectively. The Medium-load
case is especially important: LIDSA sustains LOS C at
\ensuremath{v/c=0.63}, whereas every non-LLM controller remains at LOS F.

Waiting time and mean speed follow the same pattern. LIDSA records
near-zero waiting across all loads, with 0.00 s/veh at Low load,
0.08 s/veh at Medium load, and 0.34 s/veh at High load. At Medium load,
this corresponds to reductions of 97.9\% relative to Fixed, 99.3\%
relative to SCATS, and 78.4\% relative to GLOSA. LIDSA also maintains the
highest mean network speed at every load level, most notably at Medium
load, where it reaches 40.6 km/h compared with 28.3 km/h for GLOSA and
15.5 km/h for Fixed and SCATS. This indicates that LIDSA delays the
free-flow-to-congestion transition by replacing stop-and-go control with
continuous conflict-aware speed modulation.

\subsection{Queue Suppression}
\label{subsec:queue_suppression}

\begin{figure}[!t]
  \centering
  \includegraphics[width=\columnwidth]{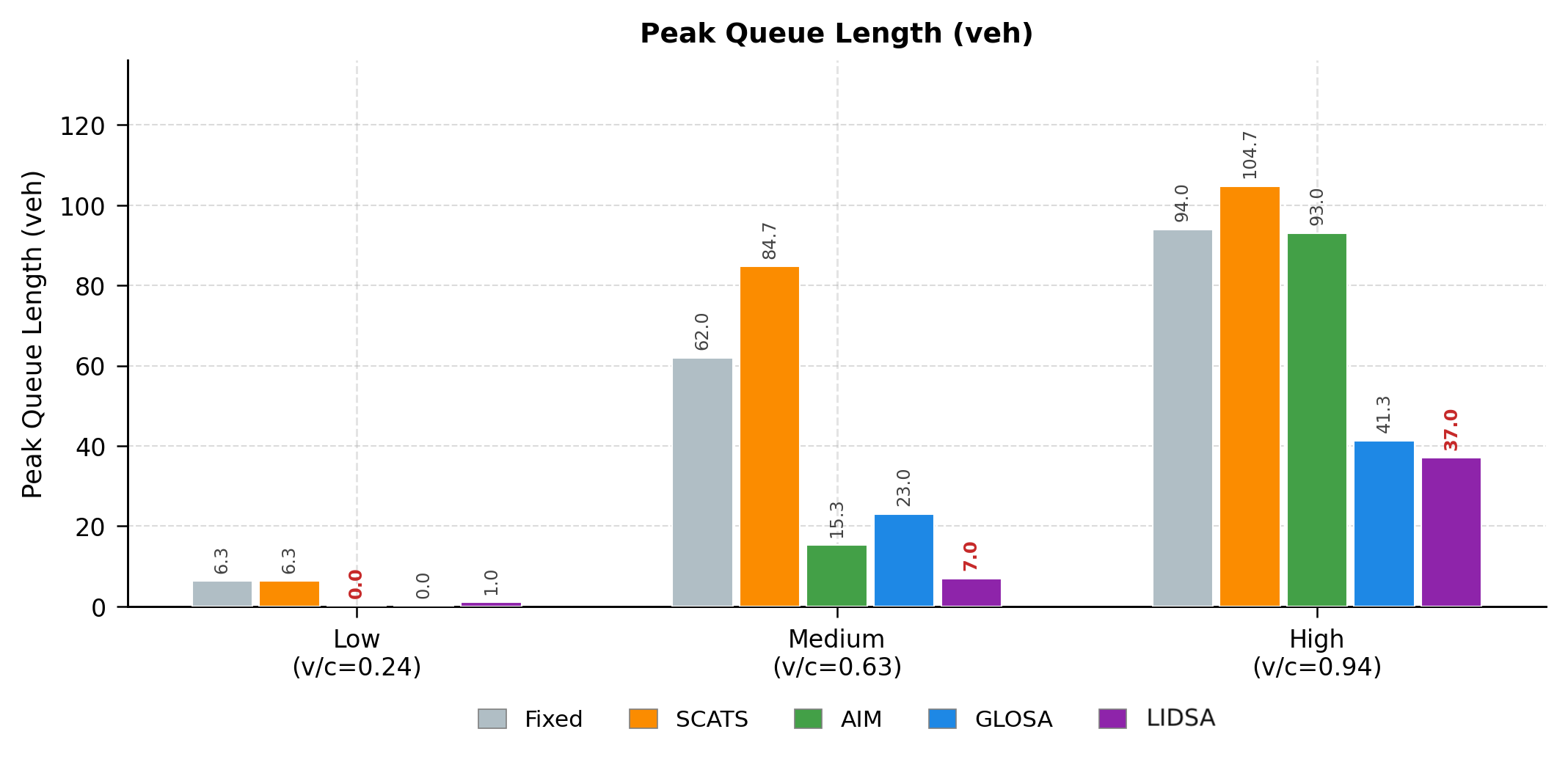}
  \caption{Peak inbound queue length (veh). }
  \label{fig:peakqueue}
\end{figure}

Queue results confirm that LIDSA reduces both average congestion and
worst-case spillback risk. At Low load, all signal-free controllers
maintain negligible queues, with LIDSA recording an average queue of
0.01 vehicles. At Medium load, LIDSA achieves the lowest average queue
length, 0.62 vehicles, compared with 1.09 vehicles for AIM, the
next-best controller. In contrast, Fixed and SCATS accumulate 38.1 and
52.9 vehicles, respectively. At High load, GLOSA records the lowest
average queue length at 8.90 vehicles, followed by LIDSA at 9.92 vehicles
and AIM at 11.0 vehicles, while both signalized controllers remain far
higher.

Peak queue length, shown in Fig.~\ref{fig:peakqueue}, highlights LIDSA's
ability to suppress transient congestion. At Medium load, LIDSA reduces
the peak queue to 7.0 vehicles, compared with 15.3 vehicles for AIM,
a 54.2\% reduction. At High load, LIDSA records the lowest peak queue
among all controllers, 37.0 vehicles, outperforming GLOSA by 10.4\%
and keeping peak queues far below the signalized baselines, both of
which exceed 94 vehicles. This behavior is consistent with the
\textsc{Share} mechanism, which prevents repeated deferral under
bilateral congestion by allowing saturated conflicting approaches to
discharge concurrently at reduced speeds.

\subsection{MAT and LLM Orchestration}
\label{subsec:mat_orchestration}
Table~\ref{tab:mat_llm_stats_flash_lite} reports the MAT and LLM
orchestration statistics for the Gemini-2.5-flash-lite backend. The
cache hit rate increases from 44.20\% under low demand to 98.83\% under
medium demand, indicating that repeated conflict signatures become highly
reusable once traffic interactions are sufficiently frequent. Under high
demand, the hit rate remains high at 87.80\%, despite a larger number of
unique states. No LLM fallbacks occurred in any scenario, and the P95
latency remained below \SI{3}{\second}, well within the anticipatory
arbitration buffer introduced by the \SI{400}{\metre} advisory horizon.

\subsection{Intent Satisfaction}
\label{subsec:intent}

\begin{figure}[!t]
  \centering
  \includegraphics[width=\columnwidth]{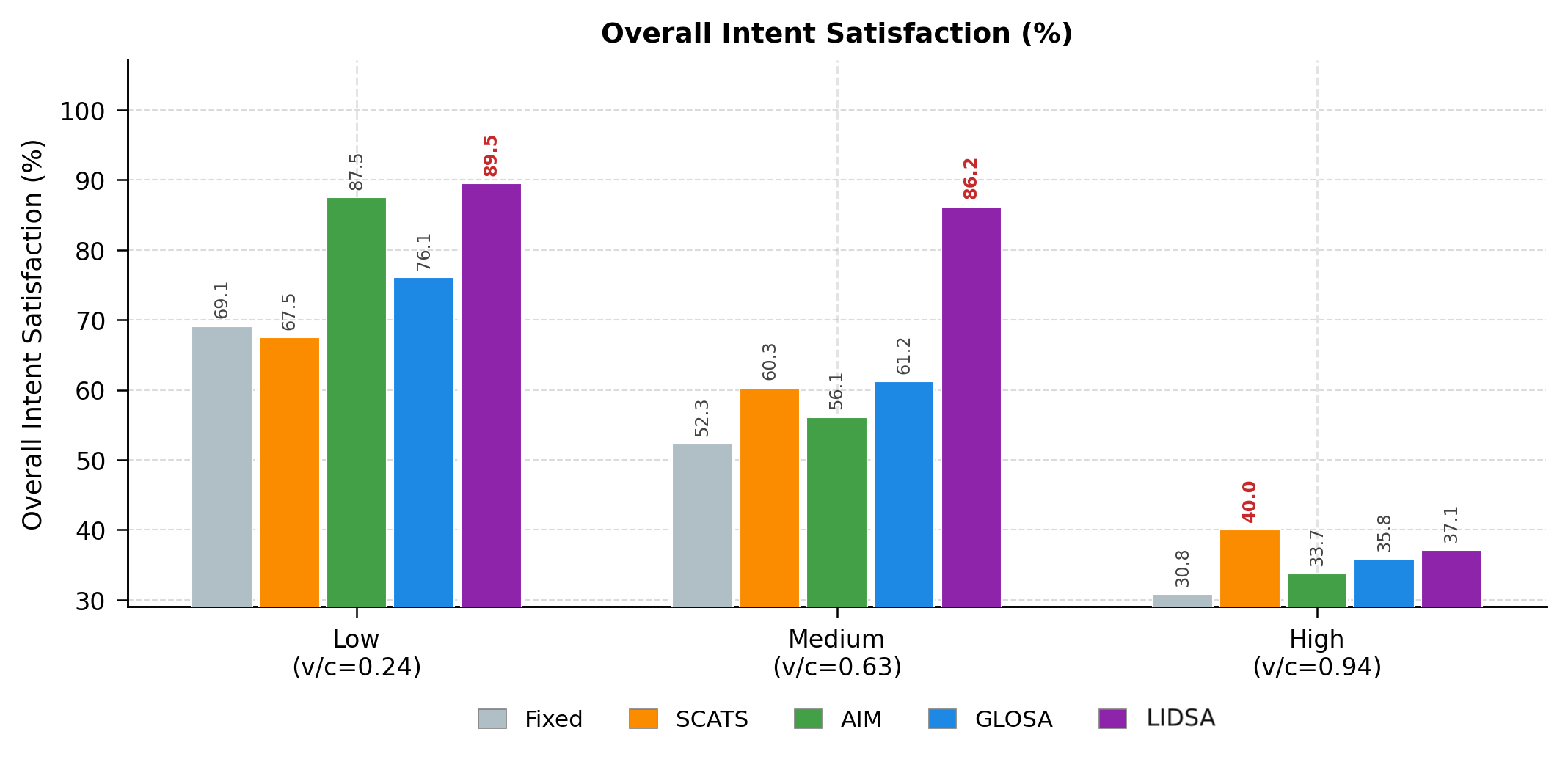}
  \caption{Overall intent satisfaction rate (\%). The score combines
           spatial, temporal, priority, and energy sub-metrics with
           weights 20/40/20/20.}
  \label{fig:intent_overall}
\end{figure}

Figure~\ref{fig:intent_overall} and Table~\ref{tab:results_summary}
show that LIDSA achieves the highest overall intent satisfaction at Low
and Medium load. At Low load, LIDSA reaches 89.5\%, compared with 87.5\%
for AIM. At Medium load, its advantage widens substantially, reaching
86.2\% compared with 61.2\% for GLOSA, the next-best controller. At High
load, all controllers are constrained by saturation; SCATS ranks first
at 40.0\%, followed by LIDSA at 37.1\%.

The overall intent result is driven mainly by temporal, priority, and
energy satisfaction. Spatial satisfaction remains 100\% for all
controllers and therefore acts as a simulation-correctness check rather
than a differentiating metric. For temporal satisfaction, LIDSA achieves
96.8\% at Low load and 95.0\% at Medium load, outperforming the
next-best controller by 4.4 percentage points and 33.7 percentage
points, respectively. Priority satisfaction follows a similar pattern:
at Medium load, LIDSA reaches 51.0\%, compared with 15.7\% for SCATS and
0\% for Fixed. LIDSA also leads energy satisfaction at all load levels,
with 96.9\%, 90.0\%, and 62.0\% under Low, Medium, and High load,
respectively. These results indicate that LIDSA's gains are not limited
to aggregate delay; they also improve the satisfaction of vehicle-level
intent objectives.

\subsection{Energy, Emissions, and Stops}
\label{subsec:energy_stops}

\begin{figure}[!t]
  \centering
  \includegraphics[width=\columnwidth]{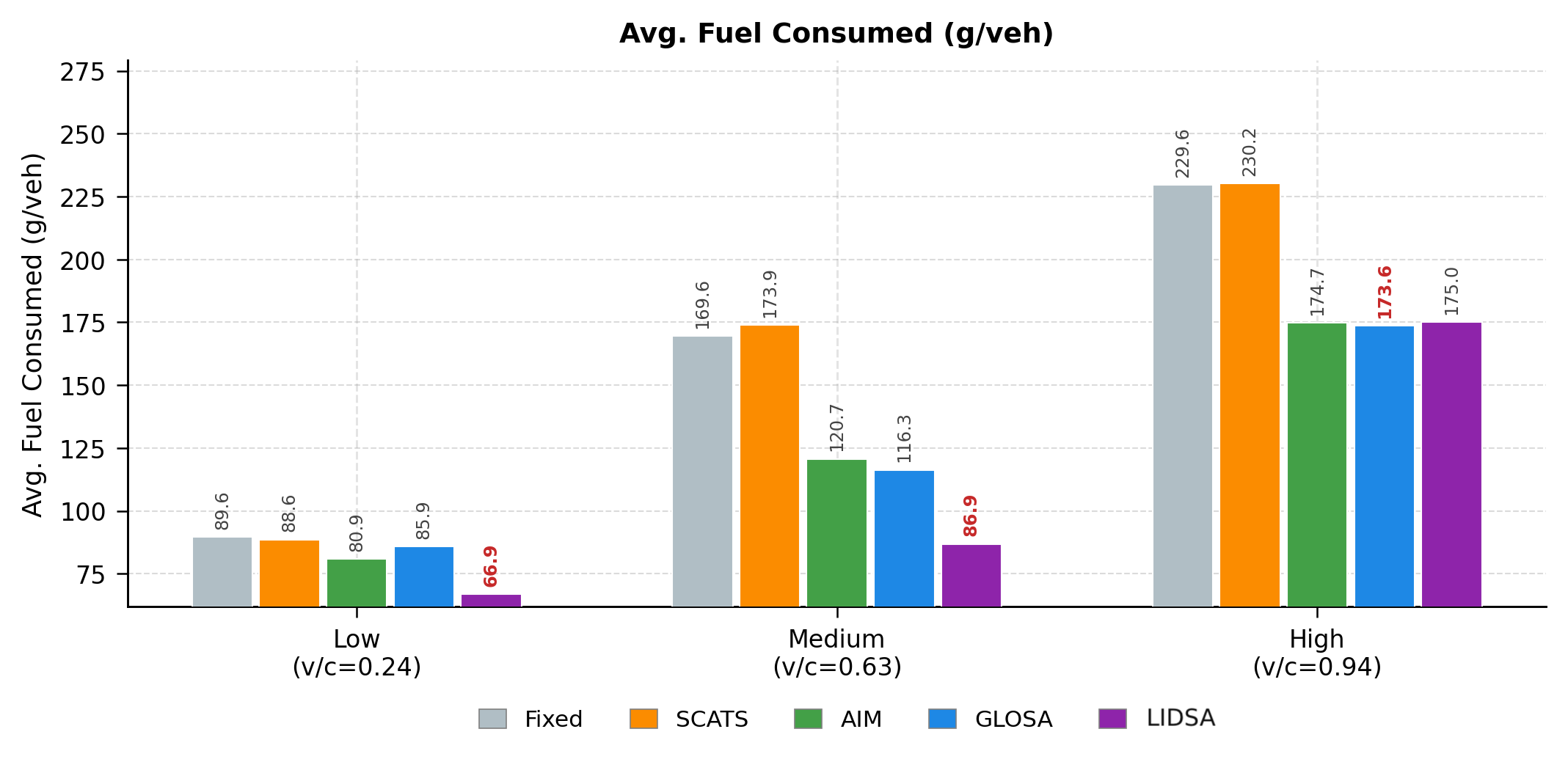}
  \caption{Average fuel consumed per vehicle (g/veh). }
  \label{fig:fuel}
\end{figure}

Fuel consumption and kinetic energy loss are estimated from each
vehicle's speed and acceleration profile, with fuel computed using the
VSP bin model described in Section~\ref{subsec:metrics}. As shown in
Fig.~\ref{fig:fuel} and Table~\ref{tab:results_summary}, LIDSA achieves
the lowest fuel consumption and kinetic energy loss at Low and Medium
load. At Low load, LIDSA consumes 66.9 g/veh and records 58.4 kJ/veh of
kinetic energy loss, reducing fuel use by 17.3\% and kinetic energy loss
by 39.2\% relative to the next-best controllers. At Medium load, LIDSA
records 86.9 g/veh and 108.9 kJ/veh, reducing fuel consumption by
25.3\% relative to GLOSA and kinetic energy loss by 19.7\% relative to
SCATS.

At High load, saturation narrows the fuel gap, with GLOSA, AIM, and
LIDSA clustered around 174--175 g/veh and all substantially below Fixed.
This suggests that LIDSA's main energy benefit occurs before full
saturation, where conflict-aware speed advisories can smooth trajectories
and reduce unnecessary braking and acceleration.

Average stops per vehicle are treated as a diagnostic smoothness metric
rather than a separate performance category. LIDSA records 1.0 stop/veh
at Low load, 1.06 stops/veh at Medium load, and 2.06 stops/veh at High
load. It performs best at Medium load, where it improves over AIM
and GLOSA and substantially outperforms the signalized controllers,
which reach 4.39 stops/veh for Fixed and 4.00 stops/veh for SCATS.
At High load, AIM and GLOSA record slightly fewer stops, but LIDSA remains
well below Fixed and SCATS. Overall, the stop-count results are
consistent with the fuel and kinetic-energy findings: LIDSA's largest
smoothness benefits occur under Low and Medium demand, before the network
enters full saturation.

\begin{table}[!t]
\centering
\caption{MAT and LLM orchestration statistics for Gemini-2.5-flash-lite with a 3600~s traffic simulation and a query cadence of \SI{30}{\second}.}
\label{tab:mat_llm_stats_flash_lite}
\footnotesize
\setlength{\tabcolsep}{3pt}
\renewcommand{\arraystretch}{1.0}
\begin{threeparttable}
\begin{tabular}{p{4.5cm}rrr}
\toprule
\textbf{Metric}
& \textbf{Low}
& \textbf{Medium}
& \textbf{High} \\
\midrule

LLM calls
& 13.00~\tnote{1}
& 18.67
& 118.33 \\

Cache hits~\tnote{2}
& 11.33
& 1584.33
& 863.67 \\

Cache misses~\tnote{3}
& 13.00
& 18.67
& 119.67 \\

Cache hit rate (\%)
& 44.20
& 98.83
& 87.80 \\

Cache size, unique states
& 13.00
& 18.67
& 118.33 \\

LLM fallbacks\tnote{4}
& 0.00
& 0.00
& 0.00 \\

Average LLM latency (ms)
& 1633.23
& 1616.00
& 1722.57 \\

P95 LLM latency (ms)
& 2603.30
& 2478.77
& 2852.33 \\

Average total tokens per call
& 3326.57
& 3429.13
& 3530.77 \\

\bottomrule
\end{tabular}

\begin{tablenotes}[flushleft]
\scriptsize
\item\tnote{1} Values are averaged over three random seeds ($\{7, 41, 129\}$).
\item\tnote{2} Cache hits denote arbitration requests served from the Memoized Arbitration Table without issuing a new LLM call.
\item\tnote{3} Cache misses denote unseen conflict signatures requiring LLM arbitration.
\item\tnote{4} LLM fallbacks denote cases where no valid response was available before timeout or output validation failed.
\end{tablenotes}
\end{threeparttable}
\end{table}

\section{Discussion}
\label{sec:discussion}

The results in Section~\ref{sec:performance} show that LIDSA can use
LLM-based arbitration to manage a signal-free intersection while
improving most evaluated efficiency, queueing, intent, and energy
metrics. Rather than treating the intersection as a fixed phase-scheduling
problem, LIDSA formulates control as context-aware right-of-way
arbitration. This section discusses the mechanisms behind these gains,
the demand regimes in which they are most pronounced, and the limitations
of the current evaluation.

\subsection{Why LIDSA Outperforms Baseline Methods}
\label{subsec:discuss_llm}

LIDSA's advantage comes from separating semantic arbitration from
low-level motion execution. The LLM receives a structured description of
approach queues, vehicle priorities, movement conflicts, and energy
preferences, and returns symbolic right-of-way roles. These roles are
then converted into deterministic speed advisories by the kinematic
executor. This separation allows LIDSA to reason jointly about conflict
avoidance, priority service, queue pressure, and trajectory smoothness
without requiring a separate hand-crafted rule for each objective.

The main practical challenge is LLM inference latency. LIDSA mitigates
this through the memoized arbitration table, which persists valid
right-of-way assignments between LLM queries. This decouples the slower
cognitive reasoning interval from the faster vehicle-advisory interval.
The high cache hit rates observed in the experiments indicate that many
traffic states remain stable enough for cached assignments to be reused,
allowing inference cost to be amortized without requiring continuous LLM
calls.

\subsection{Where Cognitive Arbitration Matters Most}
\label{subsec:discuss_medium}

LIDSA's gains are largest in the transition regime
\ensuremath{v/c=0.63}, where the intersection is neither demand-sparse
nor fully saturated. In this regime, arbitration quality strongly affects
whether vehicles clear the intersection smoothly or join a growing queue.
Fixed and adaptive signal controllers lose efficiency because phase
service is only indirectly matched to instantaneous conflict structure.
AIM resolves individual reservations but does not explicitly reason about
aggregate queue pressure. GLOSA provides speed guidance, but its
performance depends on signal-timing assumptions and predicted discharge
patterns.

LIDSA avoids these limitations by assigning right-of-way directly from the
current conflict and priority context. In particular, the
\textsc{Share} role allows compatible saturated approaches to discharge
concurrently at reduced speed, preventing repeated deferral and reducing
queue growth. This explains why LIDSA's strongest improvements appear at
Medium load, where better arbitration can still prevent the onset of
persistent congestion.

At High load, throughput converges across the best-performing
signal-free and advisory-based controllers because lane saturation flow
becomes the binding constraint. Even in this regime, LIDSA continues to
reduce control delay and peak queue formation, indicating that cognitive
arbitration remains useful for managing the distribution of congestion
even when total discharge capacity cannot be substantially increased.

\subsection{Limitations}
\label{subsec:limitations}

Several limitations bound the scope of the current evaluation and
motivate future work:

\begin{enumerate}
  \item \textit{Single-intersection topology:}
  All experiments are conducted on a single isolated four-way
  intersection. Extending LIDSA to a multi-intersection corridor will
  require coordinating arbitration across adjacent intersections to avoid
  locally optimal but corridor-level inconsistent right-of-way
  assignments. This introduces communication, architectural, and latency
  challenges that are not addressed in the present study.

  \item \textit{LLM inference latency and query cadence:}
  The 30-second query cadence was chosen to accommodate realistic
  inference latency of 1--5~s while maintaining a stable advisory
  pipeline. Although the memoized arbitration table reduces redundant LLM
  calls, the optimal cadence as a function of traffic load, intersection
  geometry, and model latency has not been systematically studied. Future
  work should evaluate adaptive query scheduling and cache-expiration
  policies.

  \item \textit{Prompt sensitivity and model generalization:}
  LLM-based controllers may be sensitive to prompt phrasing and
  output-schema design~\cite{liu2023lost}. In this study, LIDSA prompts
  were developed through iterative refinement, but they were not
  evaluated under adversarial prompt variations. Generalization to a
  broader set of open-weight models also remains untested.
\end{enumerate}

\section{Conclusion}
\label{sec:conclusion}

This paper presented LIDSA, a signal-free intersection management
framework that uses an LLM for intent-based right-of-way arbitration.
LIDSA separates high-level semantic reasoning from low-level motion
control: the LLM assigns symbolic roles, while a deterministic kinematic
executor converts those roles into per-vehicle speed advisories. A
memoized arbitration table reduces redundant inference calls, and an
independent tile-based watchdog provides a safety backstop outside the
LLM inference path.

Across the evaluated scenarios, LIDSA improves most traffic-efficiency,
queueing, intent-satisfaction, and energy metrics relative to fixed-time,
adaptive, reservation-based, and advisory-based baselines. Its strongest
gains occur in the transition regime, where arbitration quality
determines whether vehicles clear the intersection smoothly or enter
persistent queueing. Future work will extend LIDSA to multi-intersection
corridors, evaluate partial V2I penetration, and study robustness to
prompt variation and broader model families.

\bibliographystyle{IEEEtran}
\bibliography{refs}

\end{document}